\documentclass[10pt,twocolumn,letterpaper]{article}

\usepackage{cvpr}              

\usepackage{graphicx}
\usepackage{amsmath}
\usepackage{amssymb}
\usepackage{booktabs}
\usepackage{graphbox,graphicx}

\usepackage[pagebackref,breaklinks,colorlinks]{hyperref}

\usepackage[capitalize]{cleveref}
\crefname{section}{Sec.}{Secs.}
\Crefname{section}{Section}{Sections}
\Crefname{table}{Table}{Tables}
\crefname{table}{Tab.}{Tabs.}

\def\cvprPaperID{12057} 
\def\confName{CVPR}
\def\confYear{2023}

\def\eg{\emph{e.g.~}}
\def\aka{\emph{a.k.a~}}
\def\etal{{\em et al.~}}
\def\ie{\emph{i.e.~}}

\begin{document}

\title{Diverse, Difficult, and Odd Instances (D2O):\\A New Test Set for Object Classification}

\author{Ali Borji\\
Quintic AI\\
{\tt\small aliborji@gmail.com}
}
\maketitle

\begin{abstract}






Test sets are an integral part of evaluating models and gauging progress in object recognition, and more broadly in computer vision and AI. Existing test sets for object recognition, however, suffer from shortcomings such as bias towards the ImageNet characteristics and idiosyncrasies (\eg ImageNet-V2), being limited to certain types of stimuli (\eg indoor scenes in ObjectNet), and underestimating the model performance (\eg ImageNet-A). To mitigate these problems, we introduce a new test set, called D2O, which is sufficiently different from existing test sets. Images are a mix of generated images as well as images crawled from the web. They are diverse, unmodified, and representative of real-world scenarios and cause state-of-the-art models to misclassify them with high confidence. To emphasize generalization, our dataset by design does not come paired with a training set. It contains 8,060 images spread across 36 categories, out of which 29 appear in ImageNet. The best Top-1 accuracy on our dataset is around 60\% which is much lower than 91\% best Top-1 accuracy on ImageNet. We find that popular vision APIs perform very poorly in detecting objects over D2O categories such as ``faces'', ``cars'', and ``cats''. Our dataset also comes with a ``miscellaneous'' category, over which we test the image tagging models. Overall, our investigations demonstrate that the D2O test set contain a mix of images with varied levels of difficulty and is predictive of the average-case performance of models. It can challenge object recognition models for years to come and can spur more research in this fundamental area. 
Data and code are publicly available at: \href{https://drive.google.com/file/d/1oDqAStLuBGOzhbG10c1tAvsA397-YCTD/view?usp=sharing}{DATA} and 
\href{https://github.com/aliborji/D2O}{CODE}.

\end{abstract}





\vspace{-10pt}
\section{Introduction}




\begin{figure}[t]
    \vspace{-10pt}
    \centering

    \includegraphics[width=0.85\linewidth]{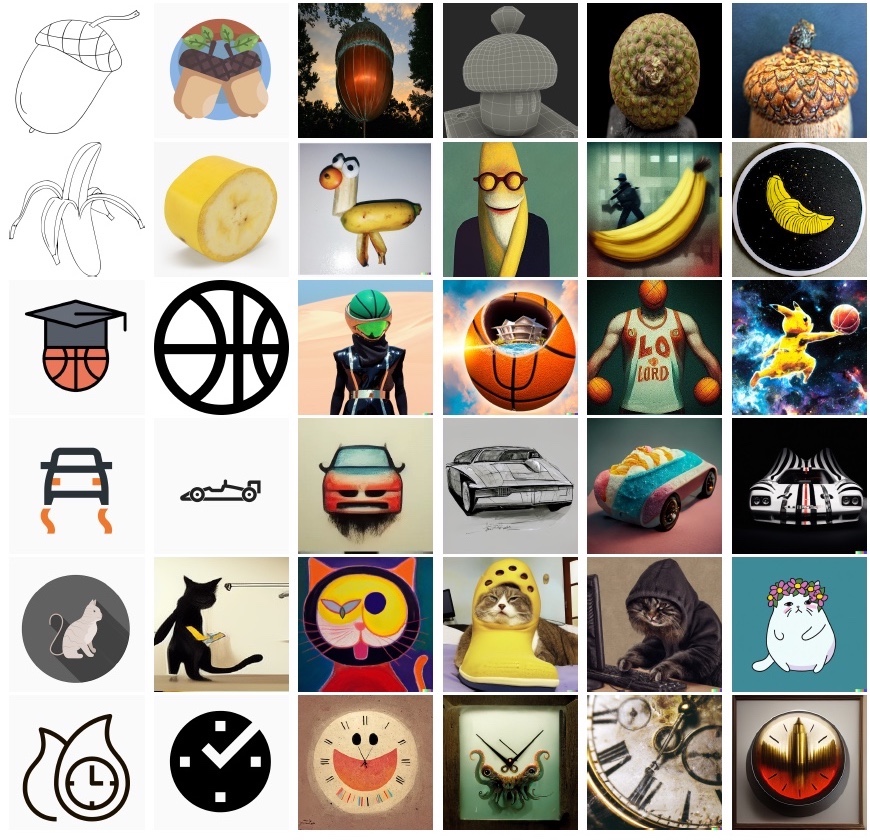}
    \caption{Sample images from D2O dataset. Images are a mix of generated images as well as images crawled from the web.
    Categories in order are: acorn, banana, basketball, car, cat, and clock.}    
    \vspace{-5pt}
    \label{fig:Samples}
    \vspace{-15pt}
\end{figure}

The object recognition problem remains in an unclear state. Despite compelling performance of state-of-the-art object recognition methods, several questions such as out-of-distribution generalization~\cite{recht2019imagenet,barbu2019objectnet,shankar2020evaluating,taori2020measuring,koh2020wilds}, ``superhuman performance''~\cite{he2015delving,geirhos2018generalisation}, adversarial vulnerability~\cite{goodfellow2014explaining}, and invariance to image transformations and distortions~\cite{hendrycks2019benchmarking} still persist. Raw performance on test sets has been the main indicator of the progress and the major feedback about the state of the field. Few test sets have been proposed for evaluating object recognition models. Some follow the footsteps of ImageNet~\cite{recht2019imagenet}. Some filter images based on failures of models~\cite{hendrycks2021natural}. Researchers have also used controlled settings to collect data~\cite{barbu2019objectnet,borji2016ilab}. While being invaluable,
these datasets suffer from few shortcomings. For example, datasets that only include examples for which the best models fail give the worst case scenario accuracy. While being useful, they underestimate model performance. Datasets that are biased towards certain environments (\eg indoor scenes in ObjectNet~\cite{barbu2019objectnet}), may not capture the full spectrum of visual stimuli. Most of the new datasets for object recognition have been centered on ImageNet (\eg ImageNet-V2, ImageNet-A, ImageNet-O) and thus may have inherited its biases. 
This may in turn give us a biased assessment of visual recognition capability of models. We argue that a good test set should strike the right balance between sample difficulty and diversity and should reflect the average-case performance of models. We also believe that new test sets that are sufficiently different from existing ones can provide new insights into the object recognition problem. To this end, we include images that contain rich semantics and require cognitive processing (\eg artistic scenes). 
Existing datasets lack enough samples of such cases. 












Here, we emphasize image diversity and difficulty over scale. While scaling up test sets has a clear advantage (\eg covering rare cases), it comes with some shortcomings. It is hard to ensure privacy, security, quality, and spurious correlations\footnote{Deep models take advantage of correlations between testing and training sets (\aka ``spurious cues'' or ``shortcuts''). These correlations are easily accessible to models but are not to humans~\cite{geirhos2020shortcut}.} in datasets containing millions of images. These problems are easier to tackle in small scale expert-made datasets. Nonetheless, both small and large datasets are needed and are complementary. Further, our dataset is one of the early efforts to use generative models for building image datasets.



Our dataset includes 8,060 images across 36 categories (Fig.~\ref{fig:Samples}). Images are carefully collected, verified, and labeled. We do not limit ourselves to object recognition models proposed in academia, and also consider prominent vision APIs in industry. This allows us to test models over a wider range of categories than those available in ImageNet and obtain a broader sense of image understanding by models. State-of-the-art models show a 30\% absolute drop in Top-1 acc on D2O test set compared to the best ImageNet accuracy (around 20\% drop using Top-5 acc). Further, over categories for which we know humans are very good at (\eg faces, cars), current APIs fail drastically.

D2O test set is intentionally not paired with a training set. It comes with a license that disallows researchers to update the parameters of any model on it. This helps avoid over-fitting on the dataset. Additionally, to mitigate the danger of leaking our data to other datasets, we mark every image by a one pixel green border which must be removed on the fly before being used.

\section{Object Recognition Test Sets}
A plethora of datasets have been proposed for image classification\footnote{Please see \href{https://paperswithcode.com/datasets?task=image-classification&page=2}{link.}}. Here, we are concerned with datasets that focus on core object recognition. ImageNet~\cite{deng2009imagenet} is one of the most used datasets in computer vision and deep learning. It contains 1,000 classes of common objects, with more than a million training images. Its test set contains 50,000 images. ImageNet test examples tend to be simple (by today's standards), clear, and close-up images of objects. As such, they may not represent harder images encountered in the real world. Further, ImageNet annotations are limited to a single label per image. To remedy the problems with ImageNet, new test sets have been proposed, which have been instrumental in gauging performance of models and measuring the gap between models and humans. The major ones are reviewed below. Several other datasets such as CIFAR-100~\cite{krizhevsky2009learning}, SUN~\cite{xiao2010sun}, Places~\cite{zhou2017places}, ImageNet-Sketch~\cite{wang2019learning}, and iLab20M~\cite{borji2016ilab} have also been introduced. 




{\bf ImageNet-V2.} Recht~\etal~\cite{recht2019imagenet} built this test by closely following the ImageNet creation process. 
They reported a performance gap of about 11\% (Top-1 acc.) between the performance of the best models on this dataset and their performance on the original test set. Engstrom~\etal~\cite{engstrom2020identifying} estimated that the accuracy drop from ImageNet to ImageNet-V2 is less than 3.6\%. Some other works have also evaluated and analyzed models on this dataset~\cite{shankar2020evaluating,taori2020measuring}.



\begin{figure}[t]
\centering

\begin{subfigure}{.5\textwidth}
  \centering
  \includegraphics[width=.9\linewidth]{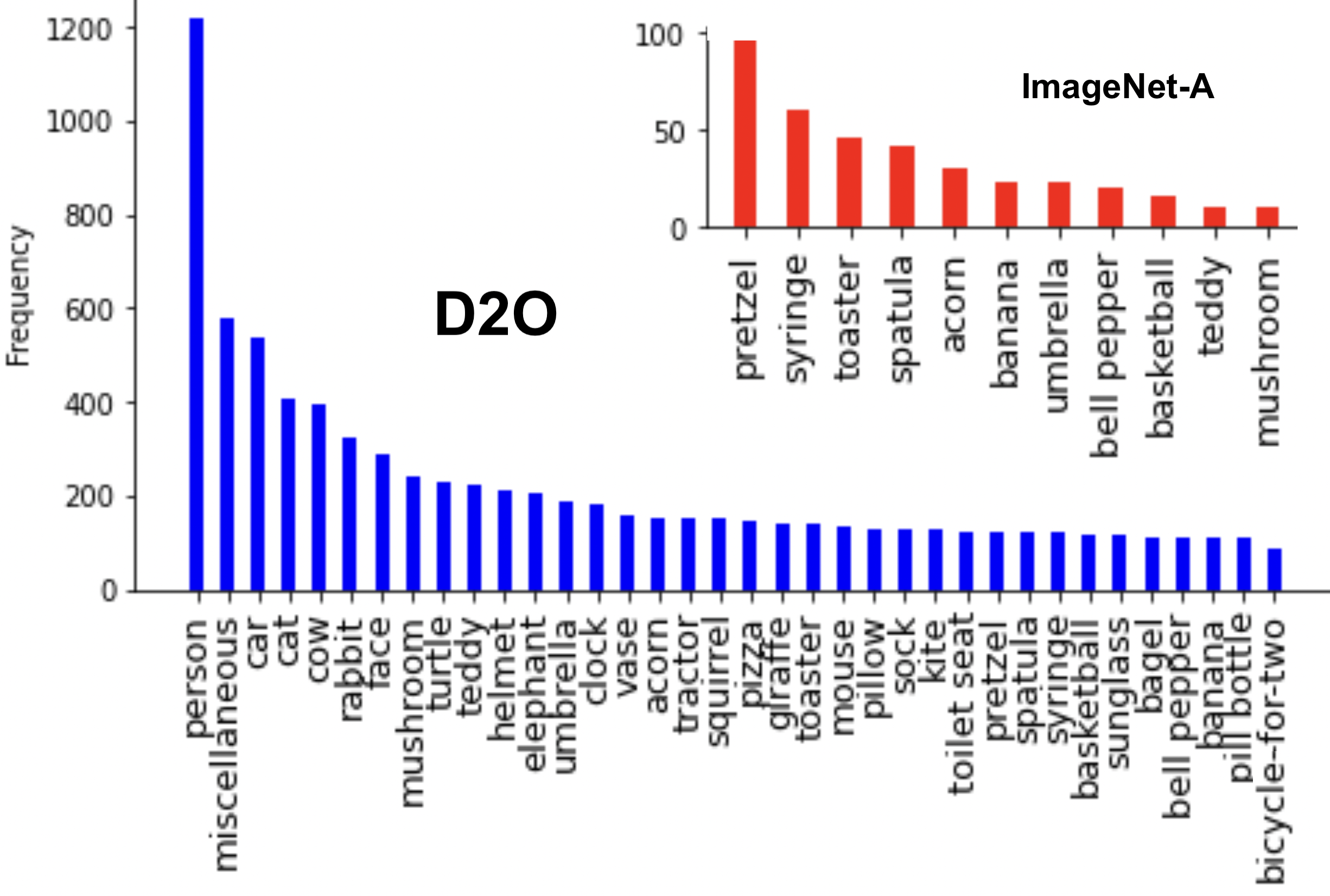} 
\end{subfigure}%
\vspace{5pt} \\
\begin{subfigure}{.45\textwidth}
  \centering
  \hspace{10pt}
    \includegraphics[width=.9\textwidth]{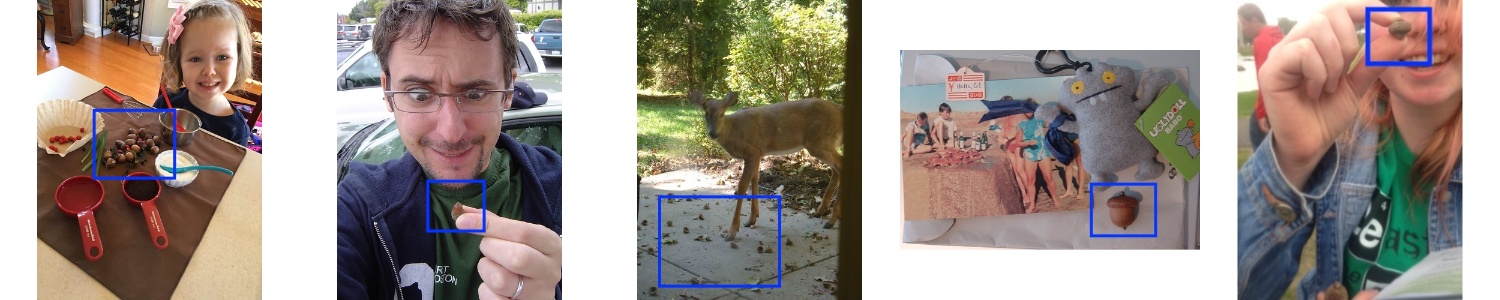} \\ 
    \hspace{20pt}
    \includegraphics[width=.9\textwidth]{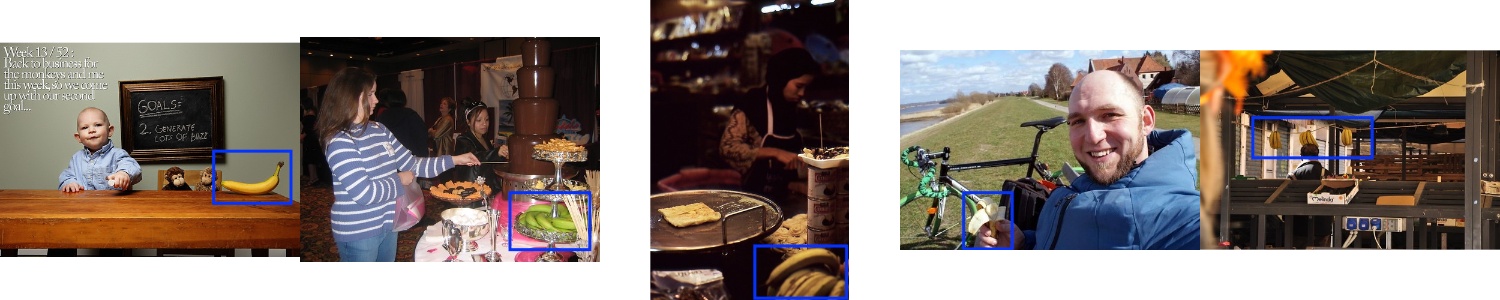} \\
    \hspace{20pt}
    \includegraphics[width=.9\textwidth]{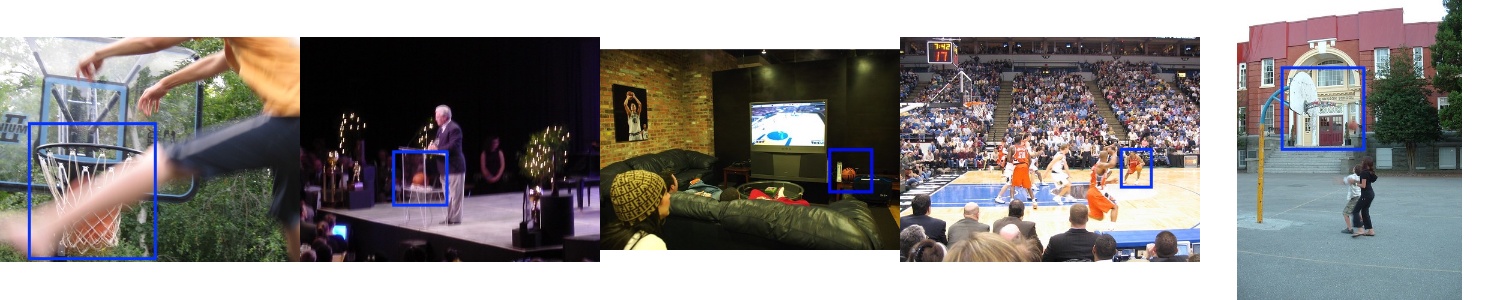}
\end{subfigure}
\vspace{-5pt}
\caption{Top: Distribution of number of objects per category over our D2O dataset as well as the 11 classes from ImageNet-A that are in common with D2O (inset panel). This subset of Imagenet-A has a total of 393 images. In total, our dataset contains 8060 images over 36 categories. Bottom: Sample images from acorn, banana, and basketball categories of Imagenet-A. The blue box demonstrates the cropped region used to build the Isolated ImageNet-A dataset. See also Appendix~\ref{appx:samples}.}
\label{fig:stats}
\vspace{-15pt}
\end{figure}

\begin{figure*}[t]
    \vspace{-10pt}
    \centering
    \includegraphics[width=.85\linewidth]{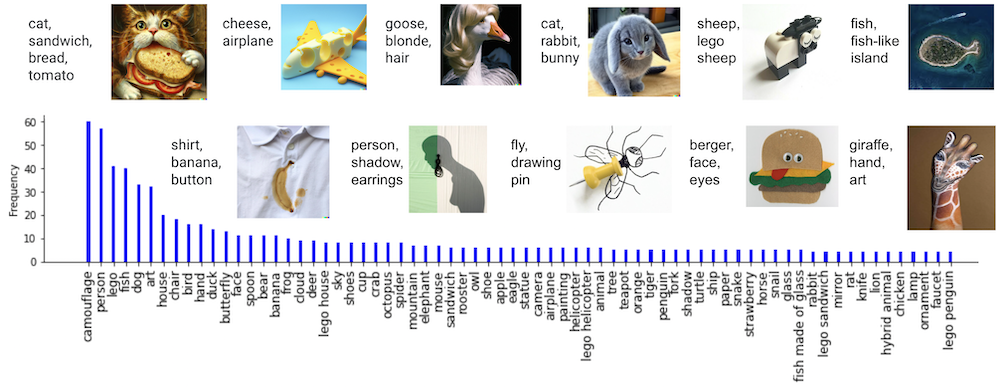} \\
    \vspace{-10pt}
    \caption{Sample images from the miscellaneous category of D2O dataset along with their tags. Bar plot shows the frequencies of the top 70 most frequent tags. Please see also Appendices~\ref{appx:modelacc} and~\ref{appx:stats} for more stats.}
    \label{fig:tags}
    \vspace{-10pt}
\end{figure*}

{\bf ImageNet-A, ImageNet-O, ImageNet-C, and ImageNet-P.} These datasets are built to measure the robustness of image classifiers against out of distribution examples and image distortions ~\cite{hendrycks2021natural,hendrycks2019benchmarking,hendrycks2020many}. Specifically, ImageNet-A dataset contains images for which a pre-trained ResNet-50 model fails to predict the correct label (Fig.~\ref{fig:stats}). It has 7,500 images scrapped from iNaturalist, Flickr, and DuckDuckGo websites. Following the approach in~\cite{hendrycks2021natural}, researchers have gathered ``natural adversarial examples'' for other problems such as object detection~\cite{lau2021natural} and microscopy analysis~\cite{pedraza2022really}. In contrast to these datasets which benchmark the worst case performance of models, here we are interested in the average case performance. To this end, instead of filtering images to fool a classifier, we include a mix of easy and hard examples to get a better sense of accuracy. Notice that a model that can only solve a very hard test set is not guaranteed to solve an easy one.

\begin{table}
\begin{center}
\begin{scriptsize}
\renewcommand{\arraystretch}{.9}
\begin{tabular}{|l|l|}
\hline
{\bf D2O class} & {\bf ImageNet class} \\
\hline \hline 
clock & digital clock, wall clock \\
elephant & Indian elephant, African elephant\\
helmet & crash helmet, football helmet, gas mask, respirator, gas helmet\\
rabbit & wood rabbit, cottontail, cottontail rabbit, Angora, Angora rabbit \\
squirrel & fox squirrel, eastern fox squirrel, Sciurus niger \\
sun glass & sun glass, sunglasses, dark glasses, shades \\
turtle & loggerhead, loggerhead turtle, Caretta caretta, \\
& leatherback turtle, leatherback, leathery turtle,  \\
& Dermochelys coriacea, mud turtle, box turtle, box tortoise\\ 

\hline 

\end{tabular}
\end{scriptsize}
\end{center}
\vspace{-10pt}
\caption{Class mapping between D2O dataset and ImageNet.}
\vspace{-15pt}
\label{tab:mapping}
\end{table}


{\bf Reassessed Labels (ReaL).} Beyer~\etal~\cite{beyer2020we} collected new human annotations over the ImageNet validation set and used them to reassess the accuracy of ImageNet classifiers. They showed that model gains are substantially smaller than those reported using the original ImageNet labels. Further, they found that ReaL labels eliminate more than half of the ImageNet labeling mistakes. This implies that they provide a superior estimate of the model accuracy.

{\bf ObjectNet.} To remove the biases of the ImageNet, Barbu~\etal ~\cite{barbu2019objectnet} introduced the ObjectNet dataset.
Images are pictured by Mechanical Turk workers using a mobile app in a variety of backgrounds, rotations, and imaging viewpoints. ObjectNet contains 50,000 images across 313 categories, out of which 113 are in common with ImageNet categories. Astonishingly, Barbu~\etal found that the state-of-the-art deep models perform drastically lower on ObjectNet compared to their performance on ImageNet (about 40-45\% drop). Later on,~\cite{borji2021contemplating} revisited the Barbu~\etal's results and found that applying deep models to the isolated objects, rather than the entire scene as is done in the original paper, leads to 20-30\% performance improvement.

\section{D2O Test Set}

We followed two approaches to collect the data. In the first one, we used publicly-available and free-to-distribute sources. We crawled images from the Flickr and Google image search engine using different search queries. The queries contained terms specifying countries, locations, materials, different times (\eg 80s), odd appearances (\eg odd helmet), etc. We also included images from various categorized panels in the search results (\eg drawing, sketch, clip art, icon, neon, clay, etc.). In the second approach, we used image generation tools such as DALL-E 2~\cite{ramesh2022hierarchical}, \href{https://www.midjourney.com/}{Midjourney}, and StableDiffusion~\cite{rombach2021highresolution} to generate some images, or searched the web for some images that are generated by these tools. We only selected the images that had good quality (\eg no dog with three eyes!). Some sample generated images are shown in Appendix~\ref{appx:generated}. We did our best to ensure that no image contains sensitive material, has high resolution, or violates the copyright law\footnote{We chose images that were public domain, did not have copyright, or were released by the government.}. The gathered images encompass a wide variety of visual concepts over both RGB images, paintings, drawings, cartoons, and clip arts. To reduce ambiguity in annotation, most of the images contain one main object. Categories were selected based on the following two criteria: a) it should be possible to collect a variety of instances for them, with different levels of difficulty, and b) one would consider model errors on them egregious (\ie confusing a cat with a dog is more troublesome than confusing a beaver with a marmot). During data collection, we emphasized choosing the odd items.

\begin{figure*}[t]
    \centering
    \vspace{-15pt}    
    \includegraphics[align=t,width=.67\linewidth]{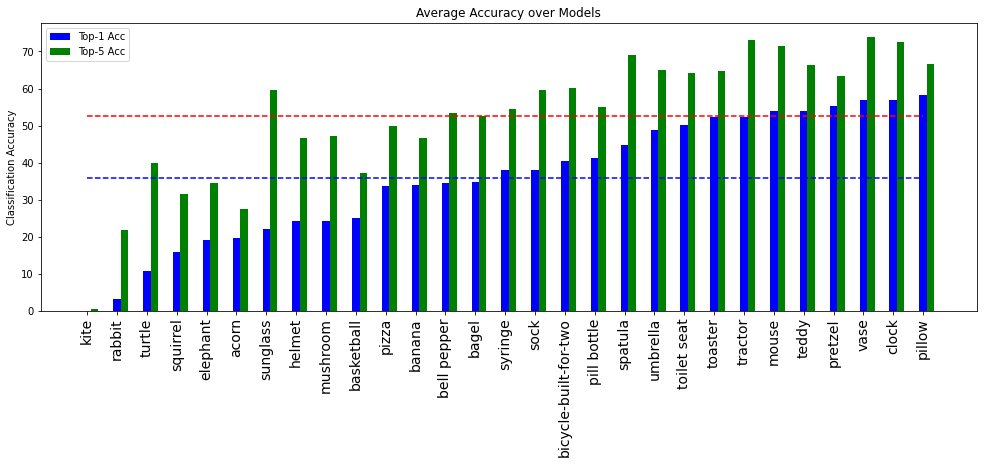}
    \includegraphics[align=t,width=.28\linewidth]{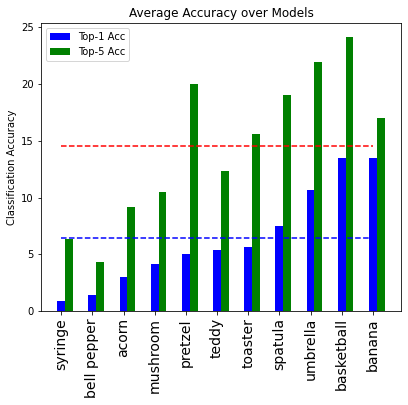}
    \vspace{-5pt}    
    \caption{Per category performance of the models on our dataset (left) and on ImageNet-A dataset (right), averaged over 10 models. The dashed lines show the average performance over categories. See Appendix~\ref{appx:modelacc} for performance of individual models.}
    \label{fig:avg_model}
    \vspace{-10pt}   
\end{figure*}

\begin{figure*}[t]
    \centering
    \includegraphics[align=t,width=.67\linewidth]{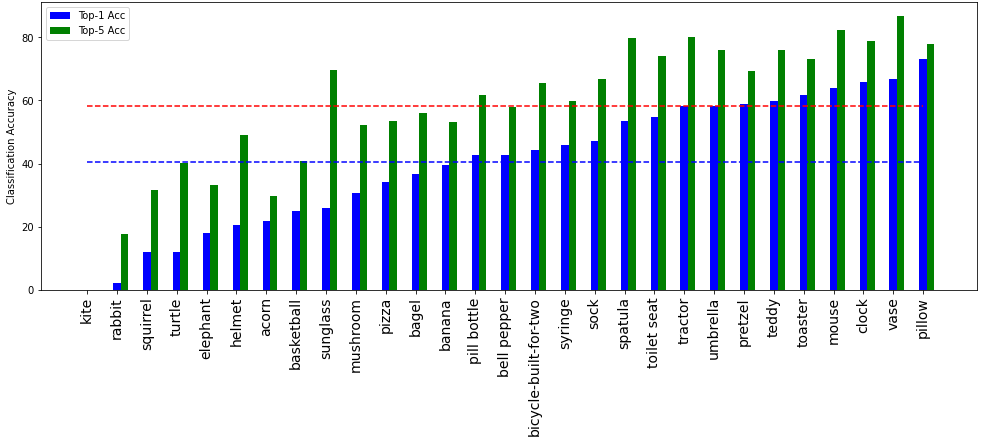}
    \includegraphics[align=t,width=.28\linewidth]{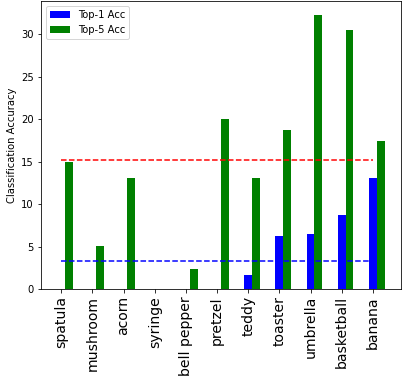}
    \vspace{-8pt}    
    \caption{Per category performance of the best model (\texttt{resnext101\_32x8d\_ws})`.}
    \label{fig:best_model}
    \vspace{-15pt}   
\end{figure*}

Three annotators were presented with an image as well as its corresponding label. They were tasked to verify the label by checking the correct or the incorrect box. The three annotators agreed with the correct label over all images.


We did not incorporate any bias towards gender, age, or race during data collection, and tried to be as inclusive as possible. Most of the categories are about objects. Few classes such as bicycle-built-for-two, face, helmet, person, sunglass, and umbrella contain humans and faces. We include and balance the number of images containing different ages and genders. The age groups are (child, 22), (teenager, 30), (adult, 51), and (elderly, 43). The gender groups include (woman, 66) and (man, 80). Notice that these issues are more important to address over large training sets. This is because sometimes models trained on such datasets are directly deployed in the real-world.

Our dataset contains 8,060 images spread across 36 categories, out of which 29 are in common with ImageNet. Six categories including \texttt{\{car, cat, cow, face, giraffe, person\}} do not appear in ImageNet, and are included mainly because they are very common and easily recognizable by humans. Seven of the categories correspond to multiple ImageNet categories, as shown in the mapping in Table~\ref{tab:mapping}. For a certain class, if a model predicted any of its corresponding ImageNet classes, we considered the prediction a hit. Sample images from our dataset are shown in Fig.~\ref{fig:Samples}. Distribution of object frequencies is shown in the top panel of Fig.~\ref{fig:stats}. The most frequent class is the \texttt{person} followed by \texttt{car} and \texttt{cat} classes. Interestingly, there is a large variation of \texttt{person} in images as this topic has fascinated many artists over time (\eg person made of wire, clay, matches, etc.). 


{\bf The miscellaneous category.} This category includes images that do not simply fall under a specific category, and may cover multiple concepts (\eg hybrid animals or strange objects). Thus, this category is suitable for testing image tagging algorithms. It has 576 images covering a wide variety of concepts and topics including hybrid animals, hybrid objects, art, illusions, camouflage objects, out of context objects, shadow, animals, fruits, drawings, paintings, objects made from different materials (\eg glass, metal, clay, cloud, tattoos, or Lego), impersonating objects, and objects from odd viewpoints. Sample images alongside their tags are shown in Fig.~\ref{fig:tags}. This figure also presents the tag frequencies. The top 10 most frequent tags are \texttt{(camouflage,60), (person,57), (Lego,41), (fish,40), (dog,33), (art,32), (house,20), (chair,18), (bird,16), (hand,16), (duck,14)}.




To put our dataset in perspective with other datasets and for cross-dataset comparison, we also evaluate models over 11 classes of the ImageNet-A dataset that also exist in our dataset. Sample images from three of these categories are shown in the bottom panel of Fig.~\ref{fig:stats}.





The D2O dataset is substantially different from ImageNet and ImageNet-A validation sets measured in terms of the Fr\'echet Inception Distance (FID)~\cite{heusel2017gans} (using 10K images). The FID between D2O and these sets in order are 45.2 and 51.3 indicating a large distribution shift, and thus high diversity. To put these numbers in perspective, the FID between ImageNet's validation and the test set is approximately 0.99. Notice that the lower the FID, the more similar the two distributions.


\section{Results and Analyses}





\subsection{Generic Object Recognition}

We tested 10 state-of-the-art object recognition models\footnote{Models are available in PyTorch hub: \url{https://pytorch.org/hub/}. We used a 12 GB NVIDIA Tesla K80 GPU to do the experiments.}, pre-trained on ImageNet, on our dataset. These models have been published over the past several years and have been immensely successful over the ImageNet benchmarks. They include AlexNet~\cite{krizhevsky2012imagenet}, MobileNetV2~\cite{sandler2018mobilenetv2}, GoogleNet~\cite{szegedy2015going}, DenseNet~\cite{huang2017densely}, ResNext~\cite{xie2017aggregated}, ResNet101 and ResNet152~\cite{resnet}, Inception\_V3~\cite{szegedy2016rethinking}, Deit~\cite{touvron2021training}, and ResNext\_WSL~\cite{mahajan2018exploring}. Details on accuracy computation are given in Appendix~\ref{appx:acc}.

Models are trained on ImageNet and tested only on the classes shared with ImageNet. Performance per D2O category, averaged over models, is shown in Fig.~\ref{fig:avg_model}. The average performance, over all models and categories, is around 30\% using Top-1 acc and around 50\% using Top-5 acc. The corresponding numbers over the ImageNet-A dataset are about 5\% and 15\%, respectively. Therefore, ImageNet-A images are on average harder than D2O images for models, perhaps because they contain a lot of clutter. 
Prior research has shown that clutter and crowding severely hinder deep models (\eg~\cite{volokitin2017deep}).
It is not clear which object is the main one in most of the ImageNet-A images (See Fig.~\ref{fig:stats}). To pinpoint whether and how much clutter contributes to low performance on this dataset, we manually cropped the object of interest in images (the blue bounding boxes in Fig.~\ref{fig:stats}). The cropped objects have low resolution, but they are still recognizable by humans. Results on this dataset, called ImageNet-A-Isolated, will be discussed in the following.

\begin{figure}[htbp]
    \centering
    \vspace{-5pt}    
    \includegraphics[width=.88\linewidth]{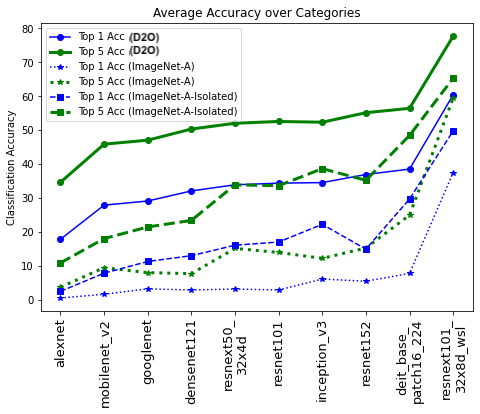}\\ 
    \includegraphics[width=.88\linewidth]{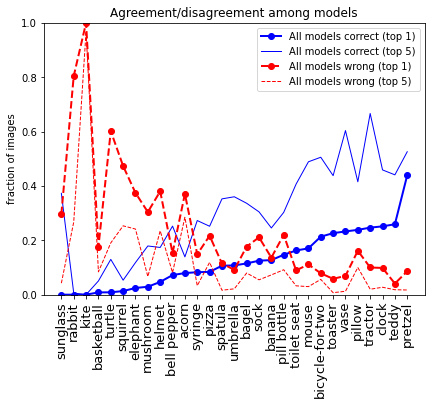}
    \vspace{-10pt}    
    \caption{\small{Top: Performance of models averaged over categories (29 categories of D2O and 11 categories of ImageNet-A). Bottom: Fraction of images over which all models fail or they all succeed.}}
    \label{fig:sum_perf}
    \vspace{-10pt}    
\end{figure}


Among the models, \texttt{resnext101\_32x8d\_ws}\footnote{This model scores 85.4\% (97.6\% top-5) over ImageNet-1k validation set (single-crop).} ranks the best over both datasets, as shown in Fig.~\ref{fig:best_model}. It achieves around 60\% Top-1 accuracy, which is much higher than its Top-1 acc over the ImageNet-A dataset ($\sim$40\%). The Top-5 acc of this model on our dataset is about 80\% compared to its 60\% over ImageNet-A. The success of this model can be attributed to the fact that it is trained to predict hashtags on billions of social media images in a weakly supervised manner. The best performance on our dataset is much lower than the best performance on the ImageNet validation set. The best Top-1 and Top-5 performance over the latter are about 91\% and 99\%, respectively\footnote{See \href{https://paperswithcode.com/sota/image-classification-on-imagenet}{paperswithcode/image-classification-on-imagenet}}. We find that better accuracy on ImageNet translates to better accuracy on D2O.

\begin{figure*}[t]
    \centering
    \vspace{-15pt}    
    \includegraphics[width=.135\linewidth]{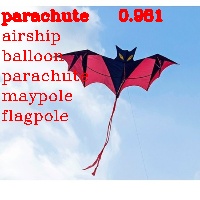}
    \includegraphics[width=.135\linewidth]{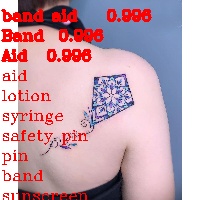}
    \includegraphics[width=.135\linewidth]{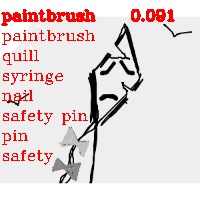}
    \includegraphics[width=.135\linewidth]{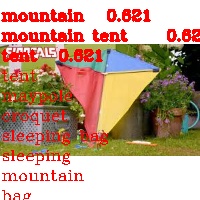}
    \includegraphics[width=.135\linewidth]{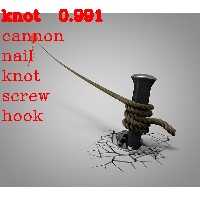}
    \includegraphics[width=.135\linewidth]{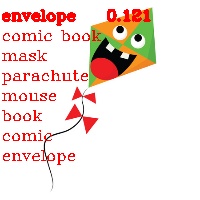}
    \includegraphics[width=.135\linewidth]{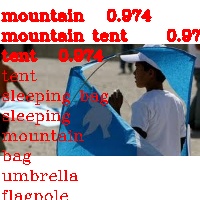}
    \rule{\textwidth}{1pt}
    \includegraphics[width=.135\linewidth]{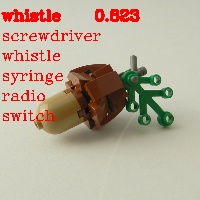}
    \includegraphics[width=.135\linewidth]{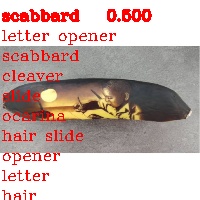}
    \includegraphics[width=.135\linewidth]{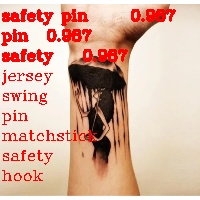}
    \includegraphics[width=.135\linewidth]{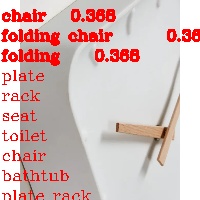}
    \includegraphics[width=.135\linewidth]{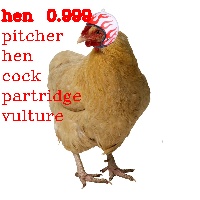}
    \includegraphics[width=.135\linewidth]{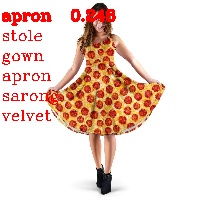}
    \includegraphics[width=.135\linewidth]{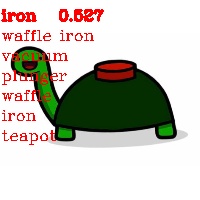}
    \includegraphics[width=.135\linewidth]{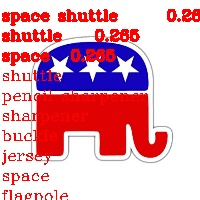}
    \includegraphics[width=.135\linewidth]{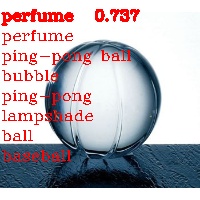}
    \includegraphics[width=.135\linewidth]{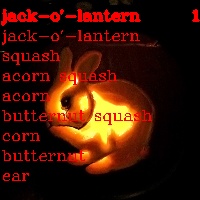}
    \includegraphics[width=.135\linewidth]{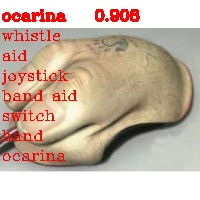}
    \includegraphics[width=.135\linewidth]{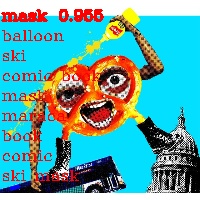}
    \includegraphics[width=.135\linewidth]{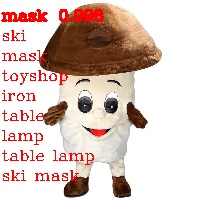}
    \includegraphics[width=.135\linewidth]{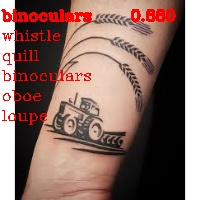}
    \caption{\small{Sample failures of \texttt{resnext101\_32x8d\_ws} model over the kite (top) and other categories (GT for bottom rows are acorn, banana, umbrella, clock, helmet, pizza, turtle, elephant, basketball, rabbit, mouse, pretzel, mushroom, and tractor). Model predictions are overlaid on images. The confidence for the top-1 prediction is also shown. Please see also Appendix~\ref{appx:filures}.}}
    \label{fig:failures}
    \vspace{-15pt}   
\end{figure*}




Performance per model, averaged over D2O categories, is shown in the top panel of Fig.~\ref{fig:sum_perf}. We observe a big difference between accuracy of \texttt{resnext101\_32x8d\_ws} model \textit{vs.} other models. This model is  $\sim$20\% better than the second best model \texttt{deit\_bas\_patch16\_224}, using Top-1 accuracy. The former model is based on weakly supervised learning whereas the latter is a transformer-based model. See Appendix~\label{appx:} for accuracy of individual models. 

As shown in Fig.~\ref{fig:sum_perf}, models perform much better over the Isolated ImageNet-A dataset than the original ImageNet-A, even though the former has low resolution images due to region cropping. This supports our argument that lower performance on ImageNet-A dataset is partly due to its scenes being cluttered. All categories enjoy an improvement in accuracy (See Appendix~\ref{appx:modelacc}).

We also test the SwinTransformer~\cite{liu2021Swin}\footnote{\url{https://github.com/microsoft/Swin-Transformer}} on D2O. This model has improved the state-of-the-art over several computer vision problems including object detection, semantic segmentation, and action recognition. 
It scores 58.2\% (Top-1) and 76.1\% (Top-5). It performs much better than ResNet, Inception and Deit models, but is slightly below the \texttt{resnext101\_32x8d\_ws} model.


\textbf{Illustrative Failure Modes.}
According to Fig.~\ref{fig:avg_model}, the top five most difficult D2O categories for all models in order are \texttt{kite, rabbit, squirrel, turtle}, and \texttt{mushroom} which happen to be the most difficult categories for the best model as well (Fig.~\ref{fig:best_model}). The kite class is often confused with parachute, balloon, and umbrella classes. Sample failure cases from the categories along with the predictions of the \texttt{resnext101\_32x8d\_ws} model are shown in Fig.~\ref{fig:failures}. 
Models often fail on drawings, unusual objects, or images where the object of interest is not unique. We also computed the fraction of images, per category, over which all models succeed, or they all fail, as shown in the bottom panel of Fig.~\ref{fig:sum_perf}. For some categories, models consistently fail (\eg kite, rabbit, turtle, squirrel), while for some others they all do very well (\eg toaster, tractor, pretzel). When all models succeed, they are correct at best over 30\% of the images (\eg toaster). This result indicates that models share similar weaknesses and strengths.




\subsection{Performance of Vision APIs }
We tested several APIs from \textbf{Microsoft}\footnote{\footnotesize{\url{https://azure.microsoft.com/en-us/services/cognitive-services/}}}, \textbf{Google}\footnote{\footnotesize{\url{https://google.github.io/mediapipe/}}}, and \textbf{MEGVII}\footnote{\footnotesize{\url{https://www.faceplusplus.com/face-detection/}}} over the D2O categories that do not exist in ImageNet: \texttt{\{face, person, car, cat, cow, giraffe\}}). These APIs are popular and highly accurate. The goal  is to see how models behave beyond ImageNet.

\begin{figure}[t]
    \centering
    \includegraphics[width=1\linewidth]{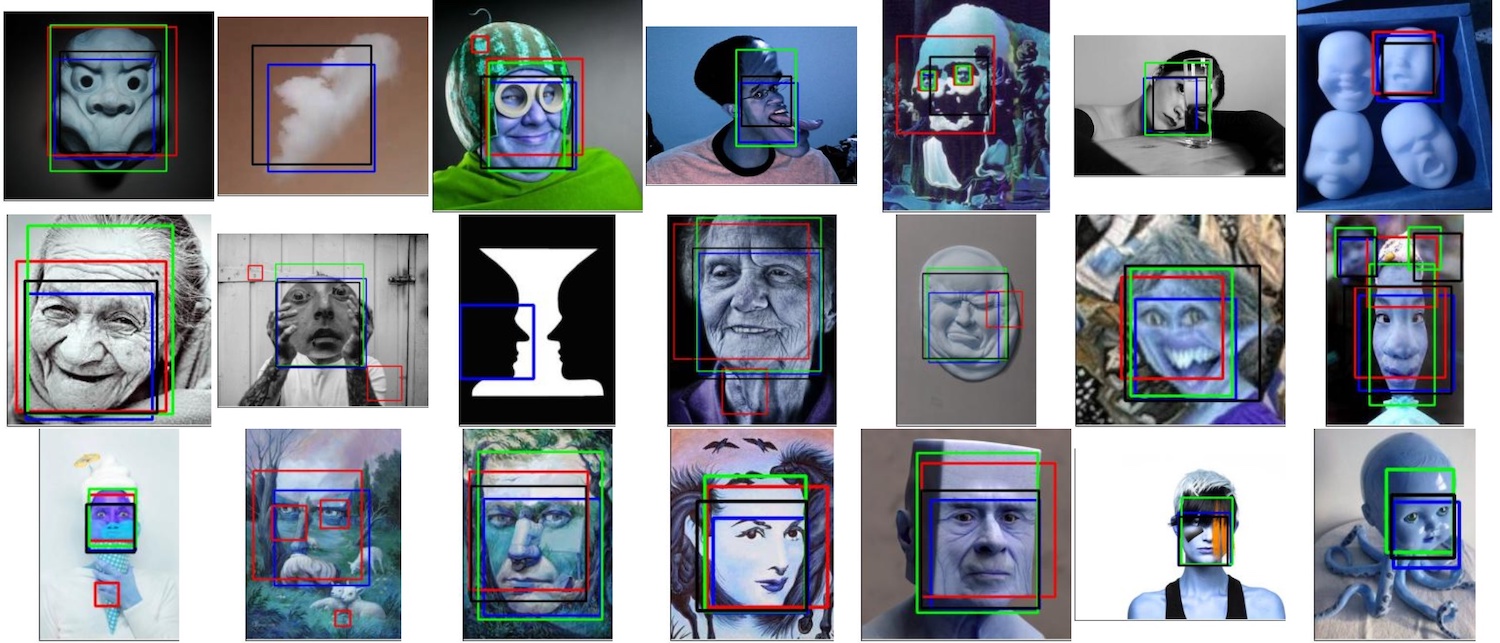}
    \caption{Sample face images along with predictions of OpenCV (Red), Microsoft API (Green), MEGVII Face++ API (Blue), and Google MediaPipe (Black) face detectors.} 
    \label{fig:faces}
    \vspace{-10pt}    
\end{figure}

\textbf{Face Detection.} 
D2O face category has 289 images and includes a lot of odd and difficult faces. Some are shown in Fig.~\ref{fig:faces}. We are mainly interested in finding whether a model is close enough in detecting faces. To this end, we refrain from using mAP to evaluate the APIs and use the accuracy score, which is easier to understand and interpret. An image is considered as a hit if the API is able to generate a bounding box with IOU greater than or equal to 0.5 with a face in the image (Ground truth boxes are annotated). Otherwise, the image is considered a mistake. We also manually verified all the predicted boxes. Our evaluation is an overestimation of performance rather than being a strict benchmark. Over the images for which the APIs failed, most often the predicted boxes did not overlap with any face in the image. In the majority of the mistakes, though, the face was missed. 

Even with the above relaxed evaluation, APIs did not do well. Microsoft Azure face detection API achieves 45.3\% accuracy in detecting D2O faces. The MEGVII Face++ API achieves 23.9\% accuracy, slightly above the 23.2\% by OpenCV face detector. Google MediaPipe face detector achieves 50.5\% accuracy. Sample face images and predictions of the APIs are shown in Fig.~\ref{fig:faces}.





\textbf{Person Detection.} MEGVII API person detector obtains about 16\% accuracy over the 1,217 person images in the person category. OpenCV person detector achieves 5\% accuracy. Microsoft object detection API achieves 27.4\% accuracy. If this API predicted a correct bounding box with any of the following classes \texttt{\{person, snowman, bronze sculpture, sculpture, doll\}}, we counted it as a hit. Evaluation is done the same way as in face detection. 





\textbf{Cat Detection.}
Over the cat category (407 images), Microsoft object detection API, predicted 95 images as cat (95/407=0.23), 9 images as Persian cat (9/407=0.022), 26 images as animal (6/407 = 0.064), and 95 images as mammal (95/407= 0.23). Considering all of these images as hits, this API achieves 54.8\% accuracy.


\textbf{Cow Detection}. 
Over the cat category (407 images), Microsoft object detection API, predicted 95 images as cat (95/407=0.23), 9 images as Persian cat (9/407=0.022), 26 images as animal (6/407 = 0.064), and 95 images as mammal (95/407= 0.23). Considering all of these images as hits, this API achieves 54.8\% accuracy.



\textbf{Giraffe Detection.}
Over the giraffe category (138 images), Microsoft object detection API predicted 30 images as giraffe (30/138=0.217), 15 images as animal (15/138=0.108), and 46 images as mammal (46/138=0.333). Considering all of these images as hits, this API achieves 65.9\% accuracy.


\textbf{Car Detection.}
Microsoft object detection API achieves 25.8\% accuracy (139 out of 539) on this category. An image was considered a hit if the API predicted a correct bounding box with any of these labels
\texttt{\{car, land vehicle, all terrain vehicle, taxi, vehicle, race car, limousine, Van, station wagon\}}.



\begin{figure}[t]
    \centering
    \includegraphics[width=\linewidth]{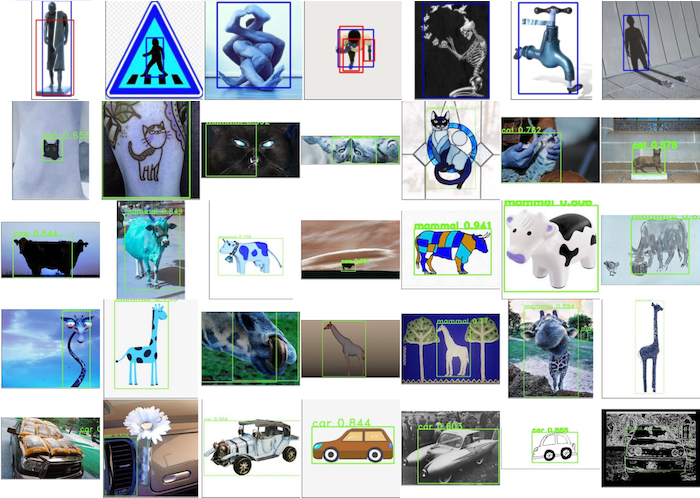}
    \caption{Sample images from the person, cat, cow, giraffe, and car categories (row-wise) along with predictions of Microsoft object detector. 
    For the person category, the blue and green boxes represent Microsoft and MEGVII person detectors, respectively.}
    \label{fig:other_objects}
\end{figure}

On the one hand, our investigation shows that APIs perform very poorly, even considering overestimated accuracy, over categories that are very easy for humans. On the other hand, it reveals that our test set is challenging for a large array of models trained on a variety of datasets. Sample images from the above categories and predictions of the APIs on them are shown in Fig.~\ref{fig:other_objects}. See also Appendix~\ref{appx:apis}.

\subsection{Tagging results}





\begin{table}
\begin{center}
\begin{footnotesize}
\renewcommand{\tabcolsep}{5pt}
\renewcommand{\arraystretch}{1}
\begin{tabular}{l|ccc|c}
{\bf API} & {\bf overlap} & {\bf no overlap} & {\bf no} & {\bf fractional}  \\
& {\bf (\% )} & {\bf (\% )} & {\bf prediction} & {\bf overlap}\\ 
\hline \hline

MSFT Tagger &  46.4  & 53.6 & 0 & 21.8 \\
\ \ \ \ " \ \ \ \ Detector &  8.5 & 36.3 &  55.2 & 17.1\\
\ \ \ \ " \ \ \ \ Recognizer &  0 & 91.3 & 8.7 & 0\\
\ \ \ \ " \ \ \ \ Captioning &  30.9 & 69.1 & 0 & 13.8\\
\hline
Google Tagger &  39.9  & 60.1 & 0 & 17.1 \\

\end{tabular}
\end{footnotesize}
\end{center}
\vspace{-10pt}
\caption{\small{Image tagging performance of APIs over the miscellaneous category of our dataset. ``No prediction'' column shows the percent of images for which there was no prediction.}}
\label{tab:api_perf}
\vspace{-10pt}
\end{table}

\begin{figure*}[t]
    \centering
    \vspace{-10pt}
    \includegraphics[width=.9\linewidth]{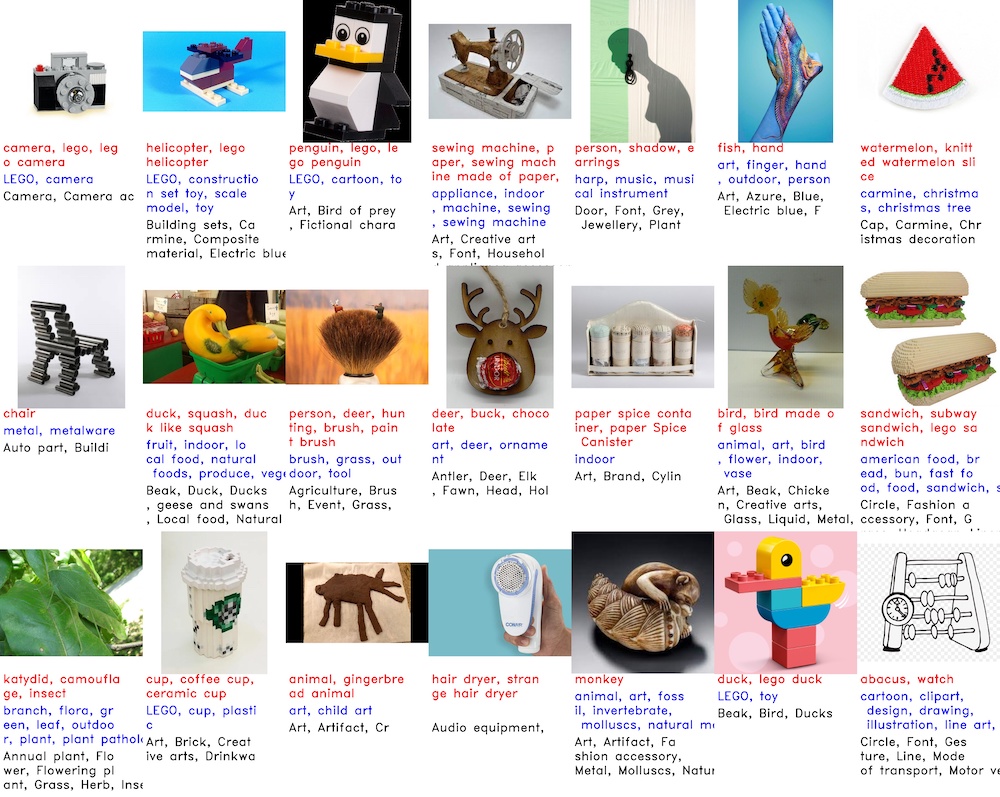}
    \vspace{-5pt}    
    \caption{{Sample images from the miscellaneous category along with ground truth tags (red) and predictions of the Microsoft (blue), and Google (black) tagging APIs.}}
    \label{fig:pred_tags}
    \vspace{-10pt}
\end{figure*}

We used the Microsoft tagging API to annotate the 576 images in the \texttt{miscellaneous} category. For 46.4\% of the images, there is a common tag between predicted tags and the ground truth tags (calculated for each image and then averaged over images). For the remaining 53.6\% of images, there is no overlap between the two sets. The fractional overlap between predicted and GT tags per image, computed as the number of common tags over the number of GT tags, is 21.8\%. The Google Vision API\footnote{\url{https://cloud.google.com/vision}} performed slightly worse than the Microsoft API. Results are shown in Table~\ref{tab:api_perf} and Fig.~\ref{fig:pred_tags}.



We also used Microsoft detection, recognition, and captioning APIs as image taggers by considering their generated labels or words as tags. Using the detection API, 
for 8.5\% of the images, there was at least one tag in common between the detected label set and the ground truth tags. For 36.3\% of the images, there was no overlap at all. For the remaining 55.2\% of images, the API did not detect anything. The fractional overlap between the predicted and ground-truth tags is 17.1\%. Using the recognition API, 91.3\% of images had no overlap and the remaining 8.7\% had no prediction at all. Using the Microsoft captioning API, for 30.9\% of the images there was an overlap between predicted and ground-truth tags (69.1\% had no overlap). The fractional overlap between the two sets is 13.8\%. Overall, the tagging APIs perform better in tagging images than APIs that are not tailored for this task, but all of them still perform poorly in tagging images. Please see Table~\ref{tab:api_perf}.





\section{Discussion and Conclusion}


We introduced a new test for object recognition and evaluated several models and APIs on it. Despite years of research and significant progress, there is still a large gap in accuracy of models on our dataset \textit{vs.} ImageNet.

Some datasets rooted in ImageNet are biased towards borrowing its images and classes, or its data collection strategy. Here, we intended to deviate from these biases. For example, unlike ImageNet-A, our dataset is model independent. ImageNet-A contains images from ImageNet for which models fail. Our dataset also includes categories, such as cows or cats, that perhaps everyone can easily recognize. These categories are missing in the ImageNet-based datasets. Further, our work encourages researchers and small teams to build carefully-curated, small-scale and versatile test sets frugally. Presently, the mindset is that datasets can only be collected by large institutions since data collection and annotation is difficult and expensive.

It is unlikely that a single test set will be enough to fully and comprehensively assess models. In practice, various test sets may be required. In conjunction with other test sets, D2O offers better insights into strengths and weaknesses of deep models. We hope that our dataset will open up new avenues for research in generalizable, robust, and human-like computer vision. We also hope that it will inspire other researchers to curate test sets where results are predictive of real-world performance. 

Our dataset has its own biases. It includes many non-natural, abstract, and artistic visualizations of objects. For a model to be able to interpret the D2O images, it has to do it similar to how humans perceive scenes and objects. 
Notice that our design decisions (\eg including certain categories such as images with boxes around objects, or not fine-tuning the models on our data) is intentional. The main goal is to use this dataset to gauge performance of deep models over time to see how general they become, rather than fitting them to solve a particular dataset. 



We share the dataset and code to facilitate future work, and will organize an annual challenge and associated workshop. 
We will share images, annotations, and metadata in a zip file. 
Our dataset is licensed under Creative Commons Attribution 4.0
(Appendix~\ref{appx:license}).

\medskip

{\small
\bibliographystyle{ieee_fullname}
\bibliography{egbib}
}

\newpage
\appendix




\section{Measuring accuracy}
\label{appx:acc}
Since some of our classes cover multiple ImageNet classes\footnote{\url{https://deeplearning.cms.waikato.ac.nz/user-guide/class-maps/IMAGENET/}}, we had to make some adjustments for computing accuracy. For example, ImageNet has three types of clocks including `digital clock', `wall clock', and `analog clock'. Here, we only have the `clock' class, containing mostly analog clocks. We chose to give the benefit of the doubt to models. A prediction is correct if the ground-truth label is in the set of the words predicted by the model. In the mentioned scenario, if a model predicts `wall clock', then a hit is counted. If the model predicts `wall' or anything else, then the prediction would be considered a mistake. The same is true for the top-5 accuracy computation. For example, if the top five model predictions are `bib', `necklace', `toilet seat', `pick', `wall clock', then the prediction is counted as a hit. In practice, first all words in the predicted labels are extracted, and then the prediction is counted as a hit if the ground-truth is in this set. In case of ground-truth having two words (\eg `toilet seat'), then it should happen in the set of words exactly as it is. 


Notice that this way of accuracy measurement gives an overestimation of the model performance, but it is good enough for our purposes here. Even with this overestimation, as we will show, models still perform poorly.



\clearpage
\newpage

\section{Samples images from the ImageNet-A and Isolated ImageNet-A datasets}
\label{appx:samples}
The rows correspond to acorn, banana, basketball, spatula, teddy, and toaster categories. The blue bounding box shows the region that we cropped to construct the Isolated ImageNet-A dataset. 

\begin{figure*}[htbp]
    \centering
    \includegraphics[width=\textwidth]{samples_imgnet_a/acorn.jpg} \\
    \includegraphics[width=\textwidth]{samples_imgnet_a/banana.jpg} \\ 
    \includegraphics[width=\textwidth]{samples_imgnet_a/basketball.jpg} \\
    \includegraphics[width=\textwidth]{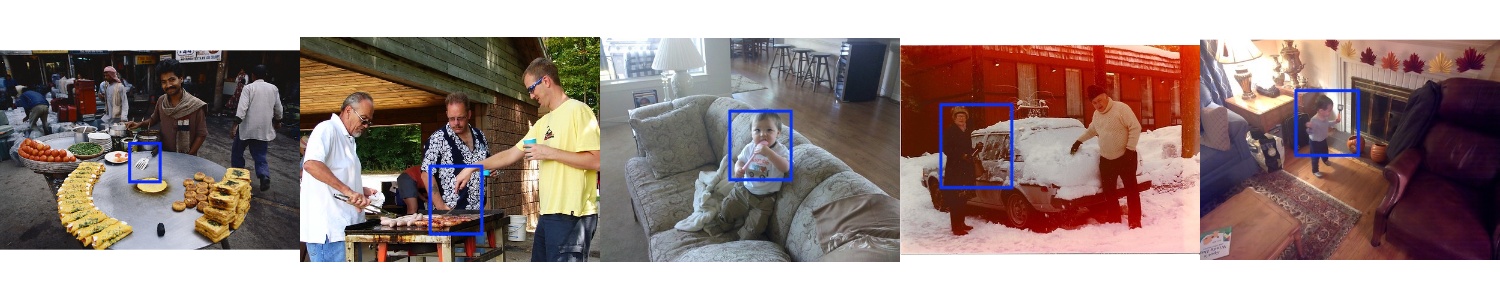}  \\
    \includegraphics[width=\textwidth]{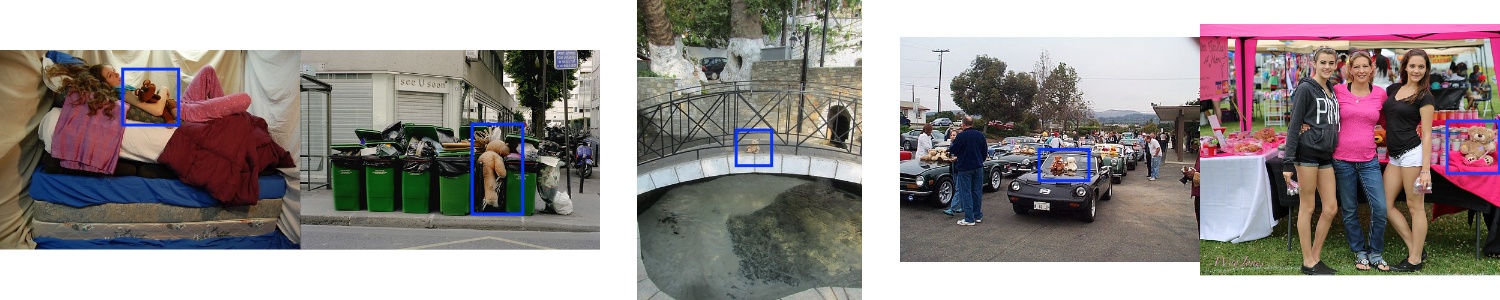} \\
    \includegraphics[width=\textwidth]{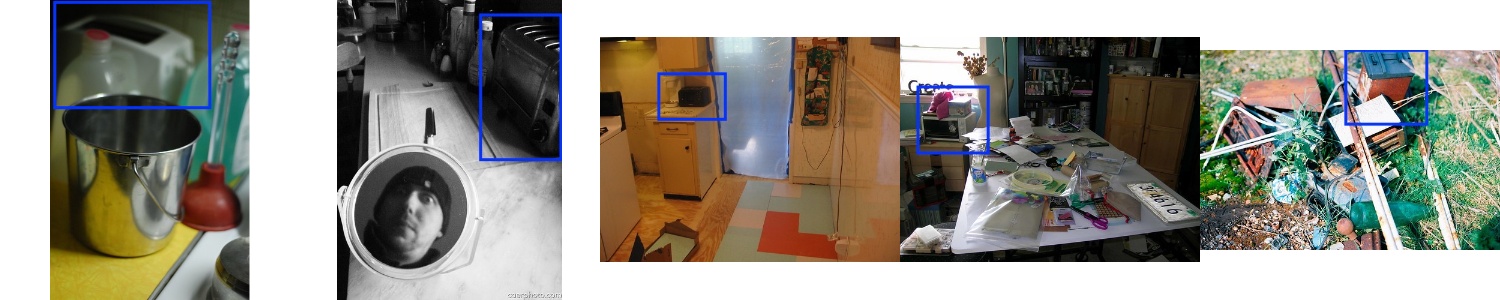}  \\
    \caption{Some sample images from the ImagenetA dataset. The blue box shows the isolated region.}
    \label{fig:appx_samples_imgnet_a}
\end{figure*}

\clearpage
\newpage
\section{Classification accuracy of individual models}
\label{appx:modelacc}


\begin{figure*}[htbp]
    \includegraphics[width=.55\linewidth]{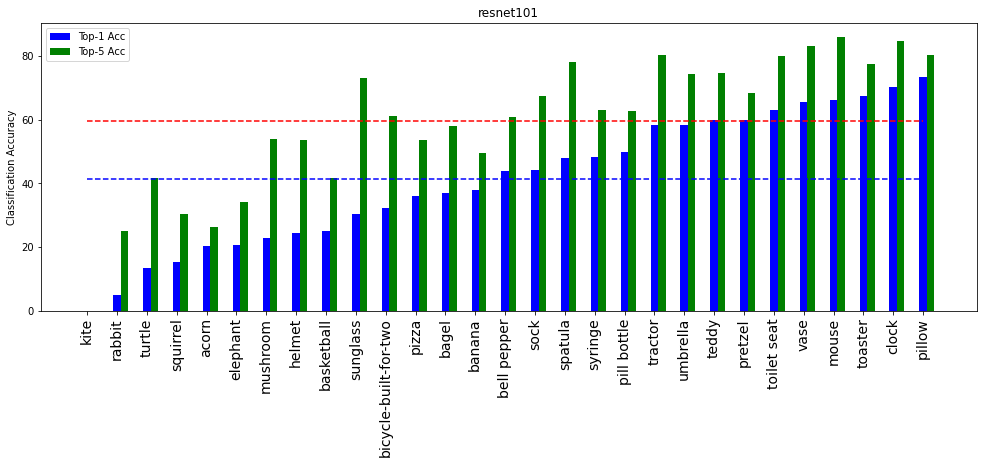} 
    \includegraphics[width=.55\linewidth]{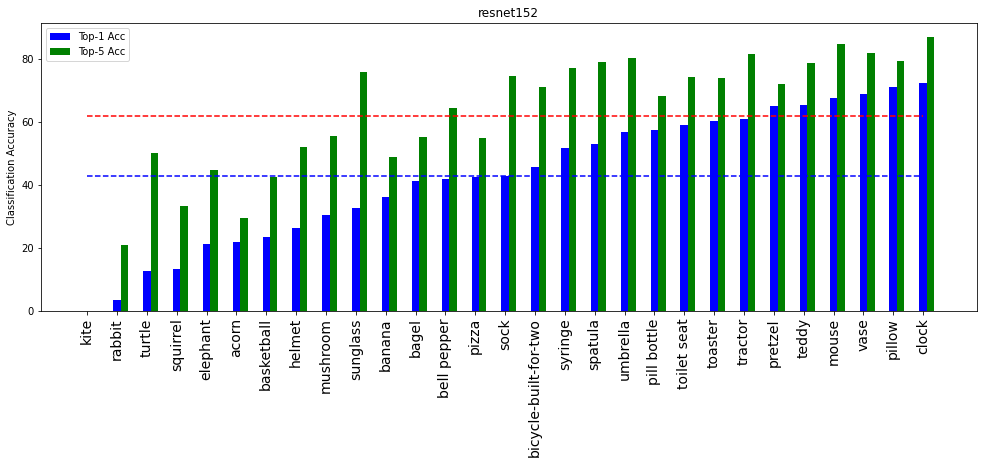} \\
    \includegraphics[width=.55\linewidth]{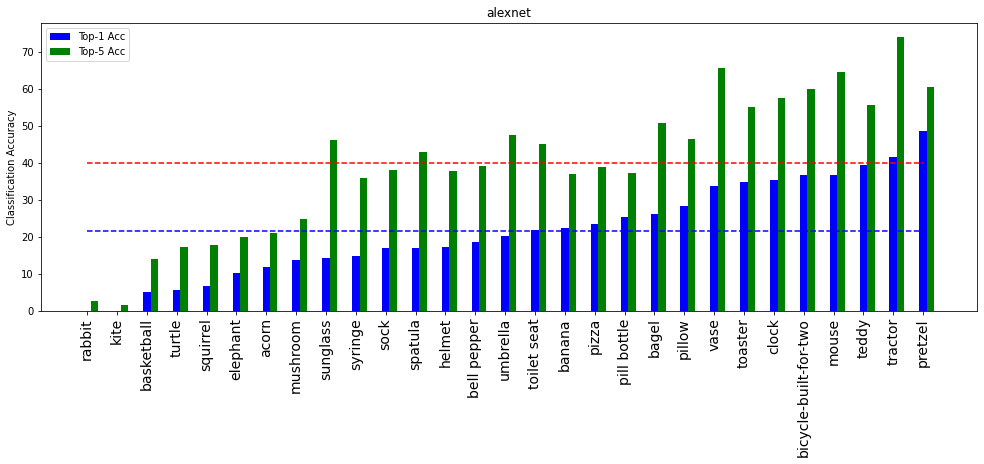} 
    \includegraphics[width=.55\linewidth]{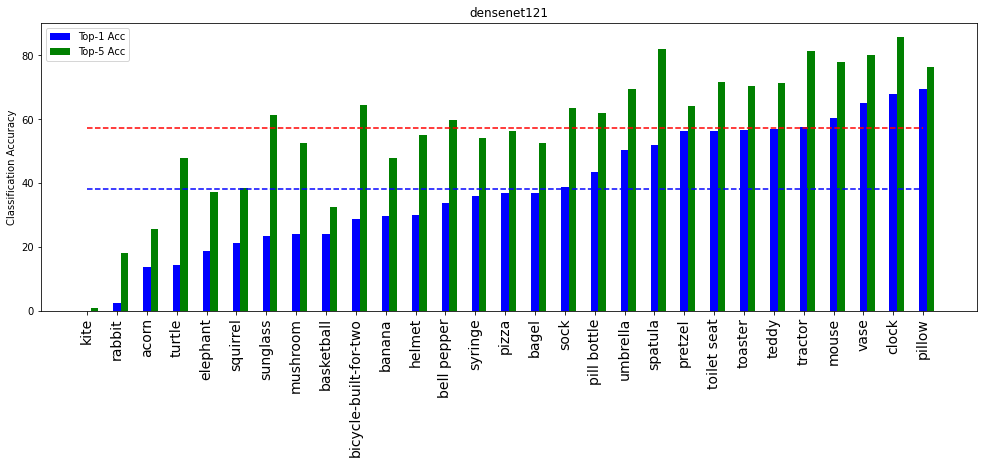} \\
    \includegraphics[width=.55\linewidth]{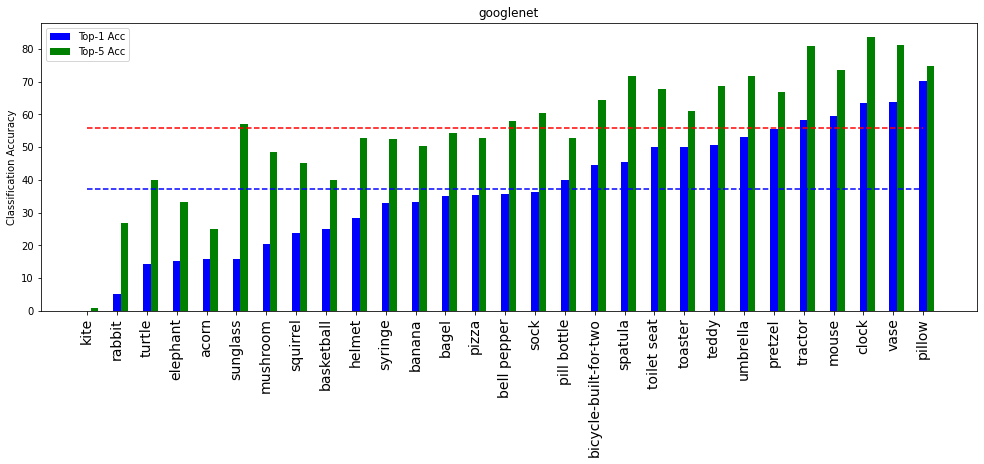} 
    \includegraphics[width=.55\linewidth]{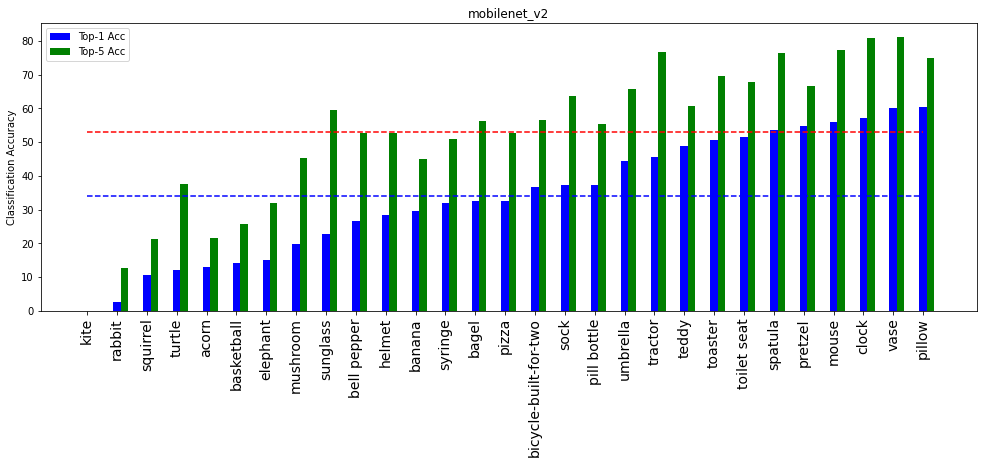} \\
    \includegraphics[width=.55\linewidth]{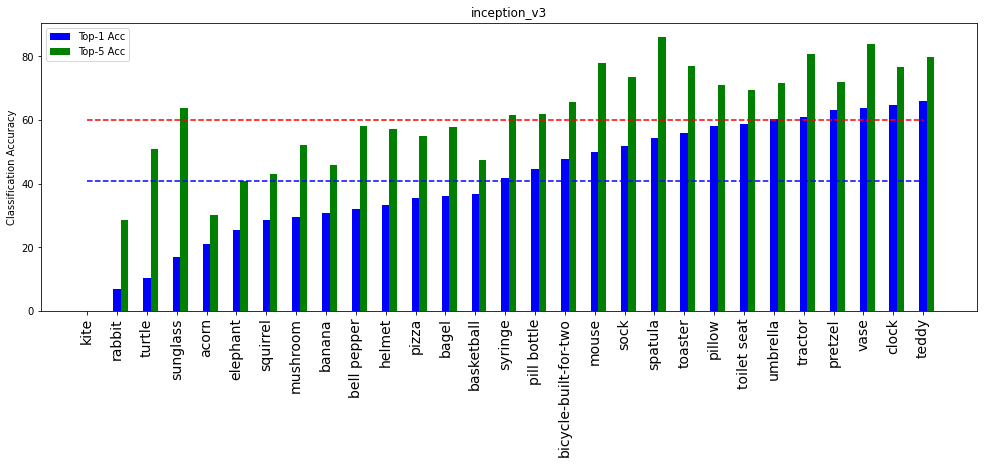} 
    \includegraphics[width=.55\linewidth]{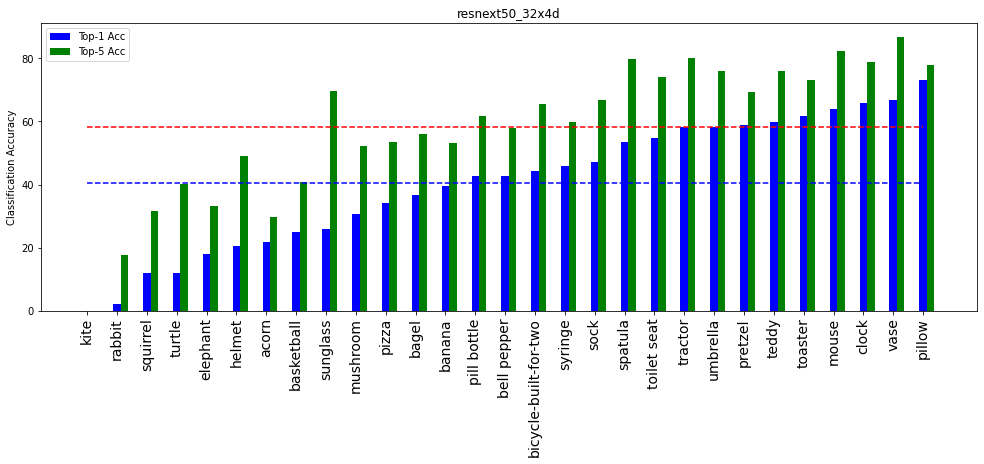} \\
    \includegraphics[width=.55\linewidth]{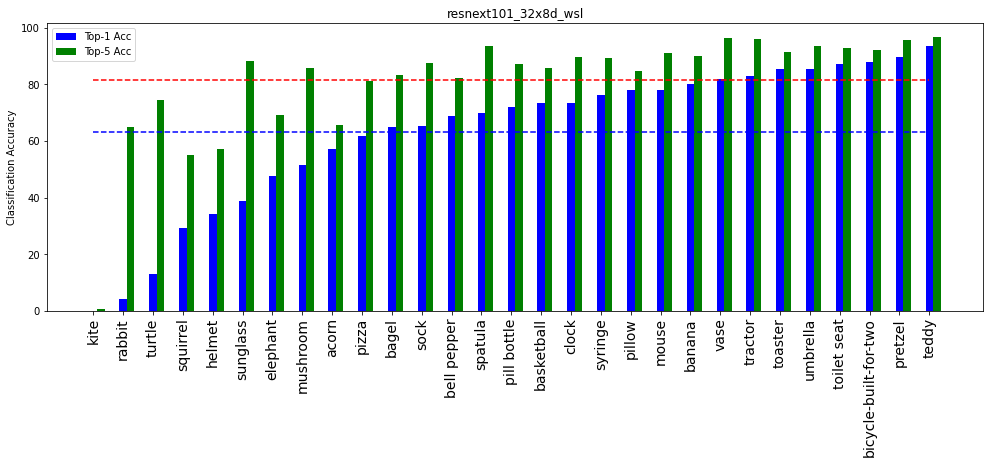} 
    \includegraphics[width=.55\linewidth]{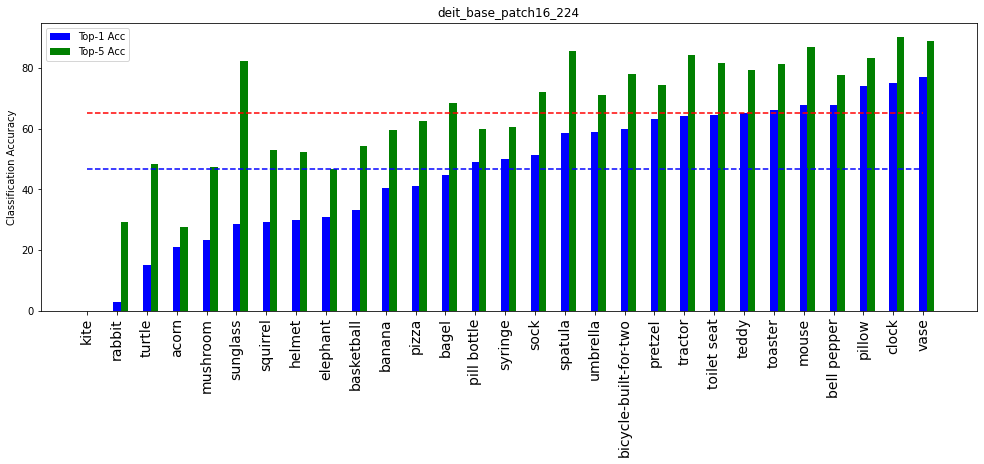} 
    \vspace{-15pt}
    \caption{Performance of models over D2O dataset.}
    \label{fig:appx_D2O_models}
\end{figure*}


\begin{figure*}[htbp]
    \centering
    \includegraphics[width=.25\linewidth]{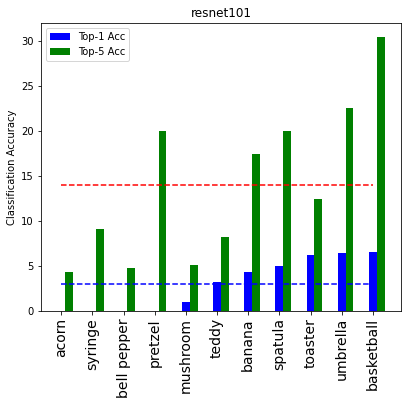} \hspace{20pt}
    \includegraphics[width=.25\linewidth]{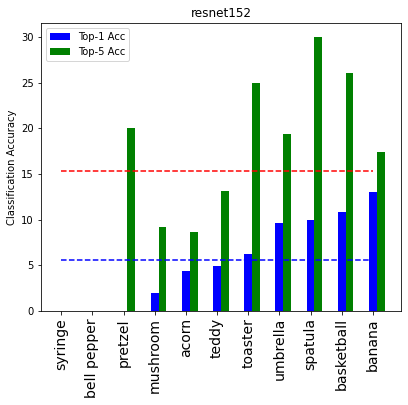} \\
    \includegraphics[width=.25\linewidth]{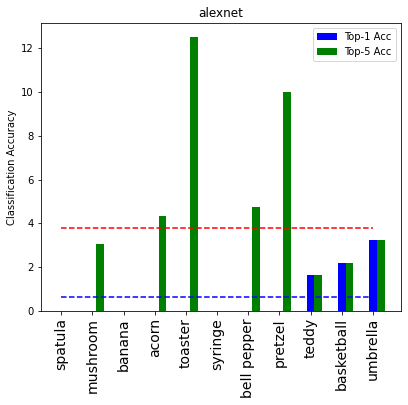} \hspace{20pt}
    \includegraphics[width=.25\linewidth]{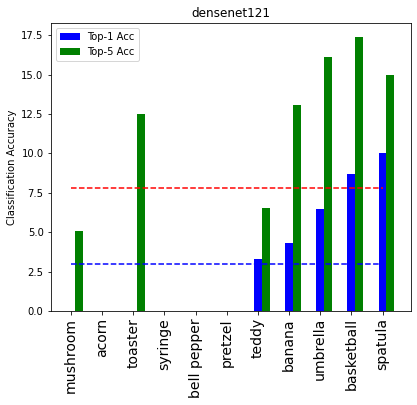} \\
    \includegraphics[width=.25\linewidth]{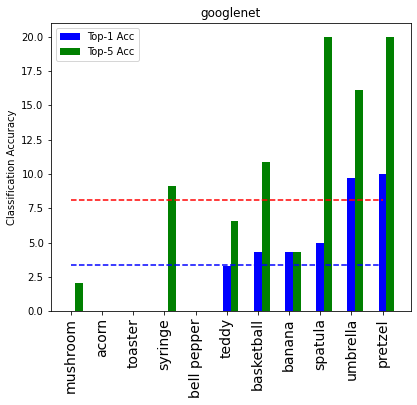} \hspace{20pt}
    \includegraphics[width=.25\linewidth]{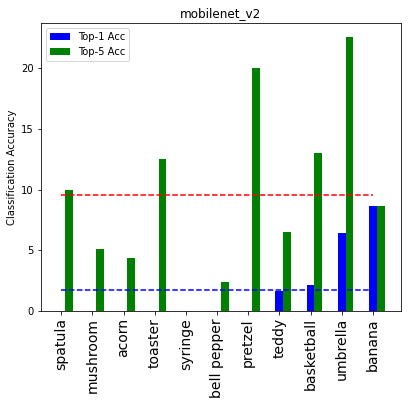} \\
    \includegraphics[width=.25\linewidth]{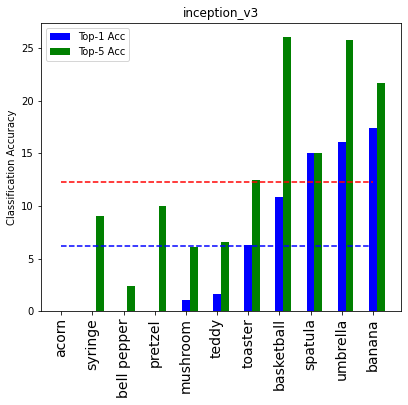} \hspace{20pt}
    \includegraphics[width=.25\linewidth]{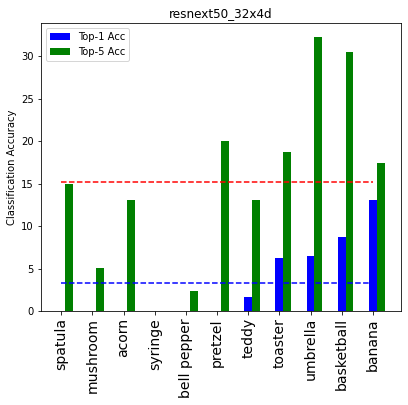} \\
    \includegraphics[width=.25\linewidth]{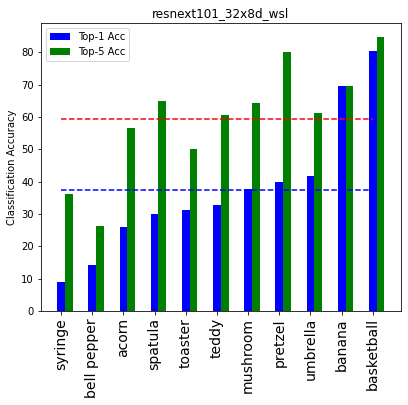} \hspace{20pt}
    \includegraphics[width=.3\linewidth]{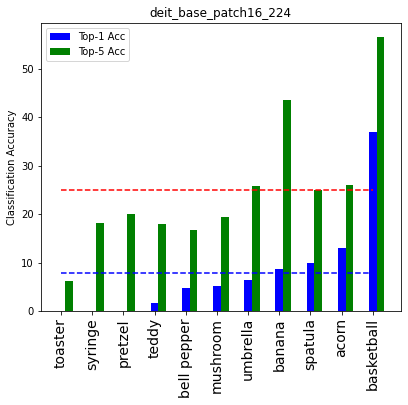} 
    \vspace{-5pt}
    \caption{Performance of models over ImageNet-A dataset.}
    \label{fig:appx_imgnet_a_models}
\end{figure*}

\begin{figure*}[htbp]
    \centering
    \includegraphics[width=.25\linewidth]{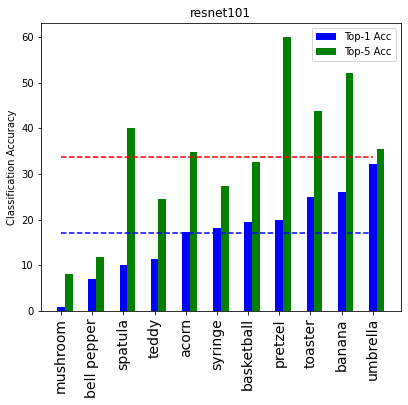} \hspace{20pt}
    \includegraphics[width=.25\linewidth]{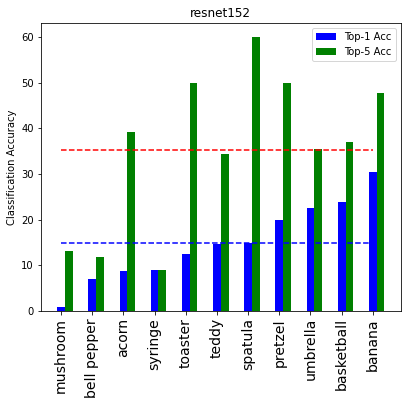} \\
    \includegraphics[width=.25\linewidth]{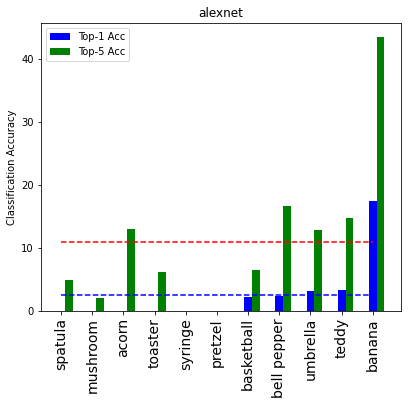} \hspace{20pt}
    \includegraphics[width=.25\linewidth]{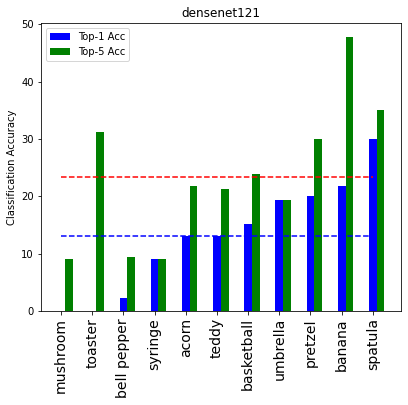} \\
    \includegraphics[width=.25\linewidth]{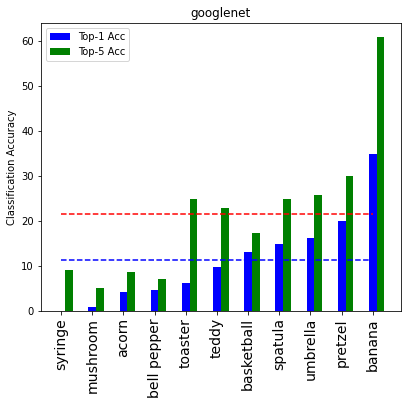} \hspace{20pt}
    \includegraphics[width=.25\linewidth]{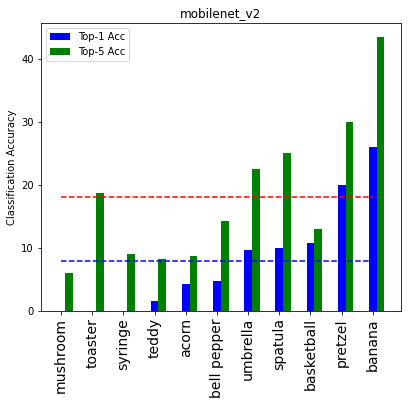} \\
    \includegraphics[width=.25\linewidth]{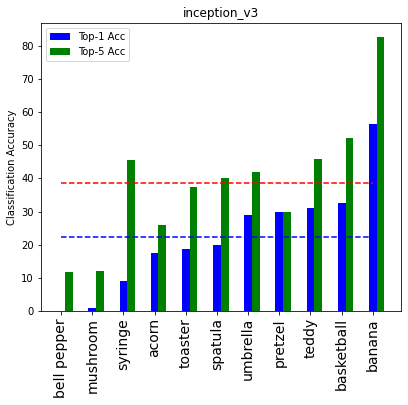} \hspace{20pt}
    \includegraphics[width=.25\linewidth]{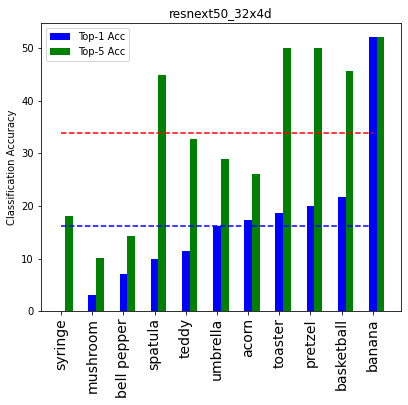} \\
    \includegraphics[width=.25\linewidth]{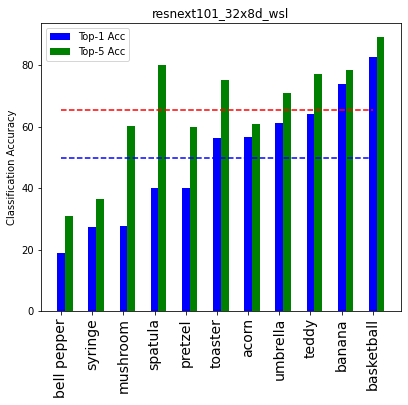} \hspace{20pt}
    \includegraphics[width=.25\linewidth]{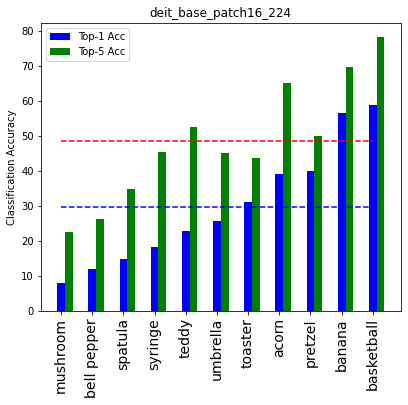} 
    \vspace{-5pt}
    \caption{Performance of models over Isolated ImageNet-A dataset.}
    \label{fig:appx_isolated_imgnet_a_models}
\end{figure*}

\clearpage
\newpage
\section{Frequency of ground-truth tags and model predicted tags}
\label{appx:stats}

\begin{figure*}[htbp]
\centering
\vspace{-10pt}
    \includegraphics[width=.7\linewidth]{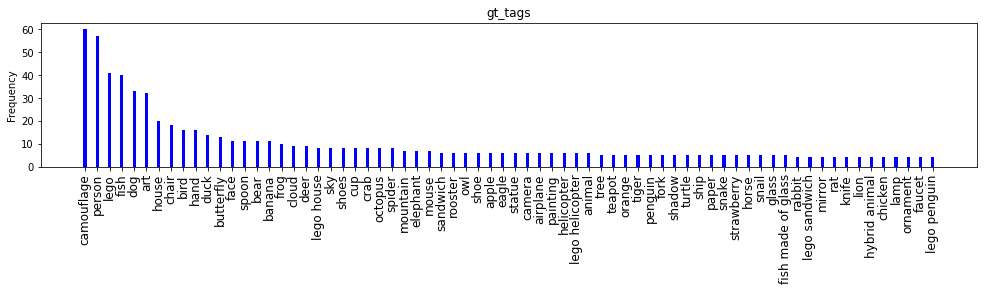} \\    %
    \includegraphics[width=.7\linewidth]{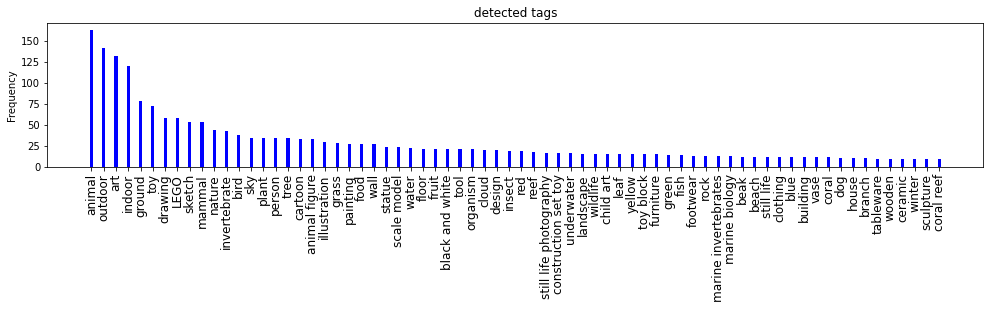} \\
    \includegraphics[width=.7\linewidth]{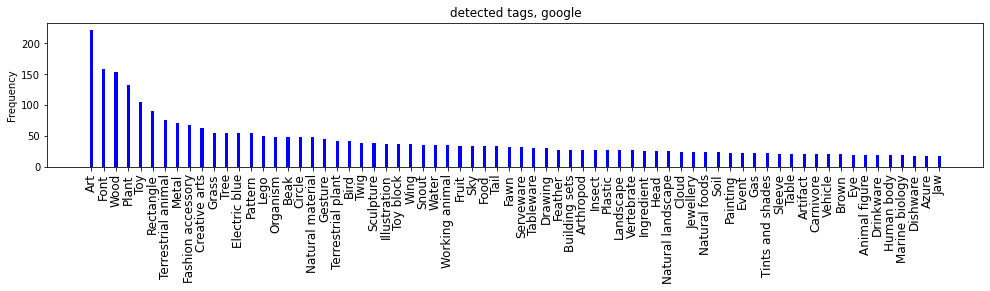} \\
    \includegraphics[width=.7\linewidth]{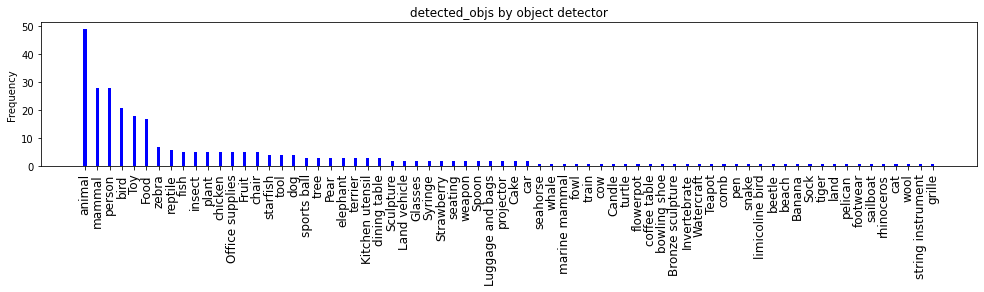}     \\
    \includegraphics[width=.7\linewidth]{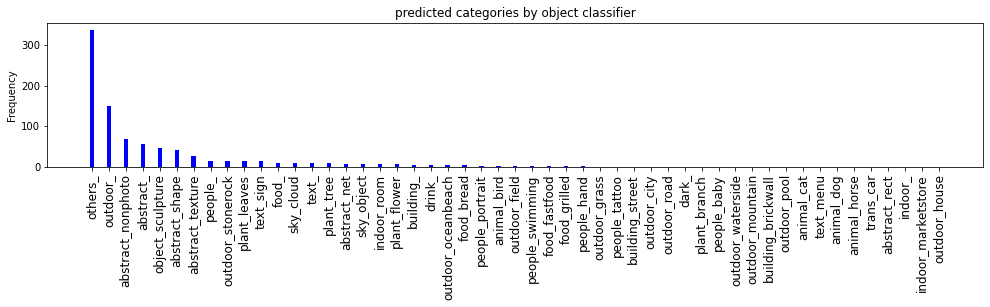} \\    
    \includegraphics[width=.7\linewidth]{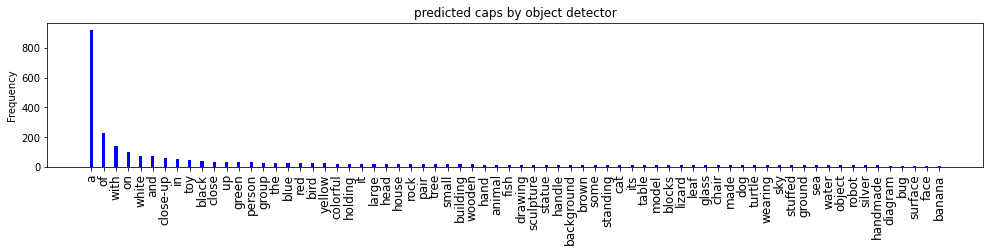} 
    \caption{\small{Frequency of ground-truth tags and model predicted tags. From top to bottom: gt tags and predicted tags by a) Microsoft API, b) Google API, c) MSFT object detector, d) MSFT object classifier, and e) MSFT image captioning.}}
    \label{fig:appx_tags}
    \vspace{-100pt}
\end{figure*}

\clearpage
\newpage
\section{Sample failure cases of best model (\texttt{resnext101\_32x8d\_ws}) over D2O dataset}


\begin{figure*}[htbp]
\centering 
 \includegraphics[width=1\linewidth]{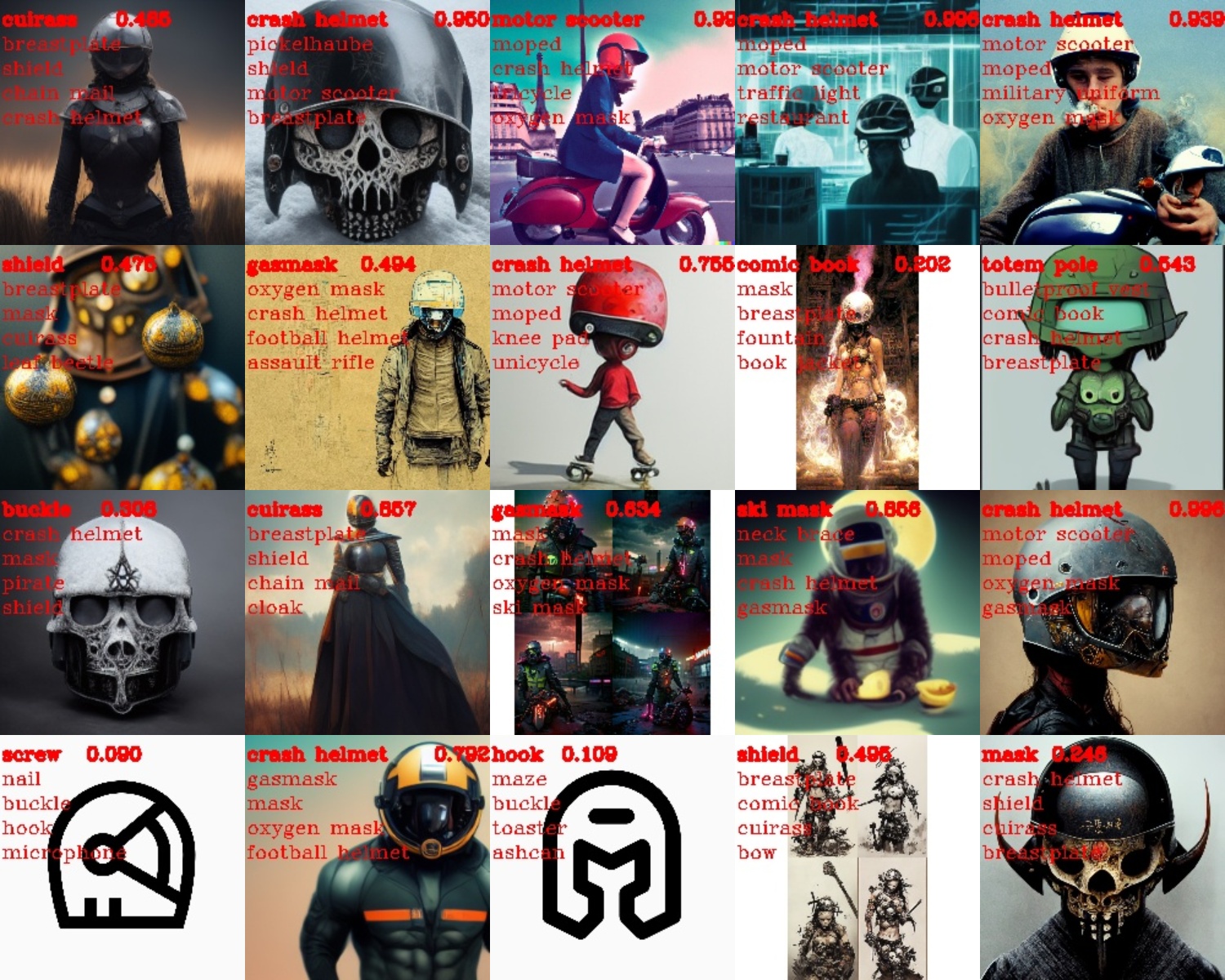}
\caption{Predictions of the resnext101\_32x8d\_wsl model over helmet category.} 
\label{fig:Errors_helmet}
\end{figure*} 

\begin{figure*}[htbp]
\centering 
 \includegraphics[width=1\linewidth]{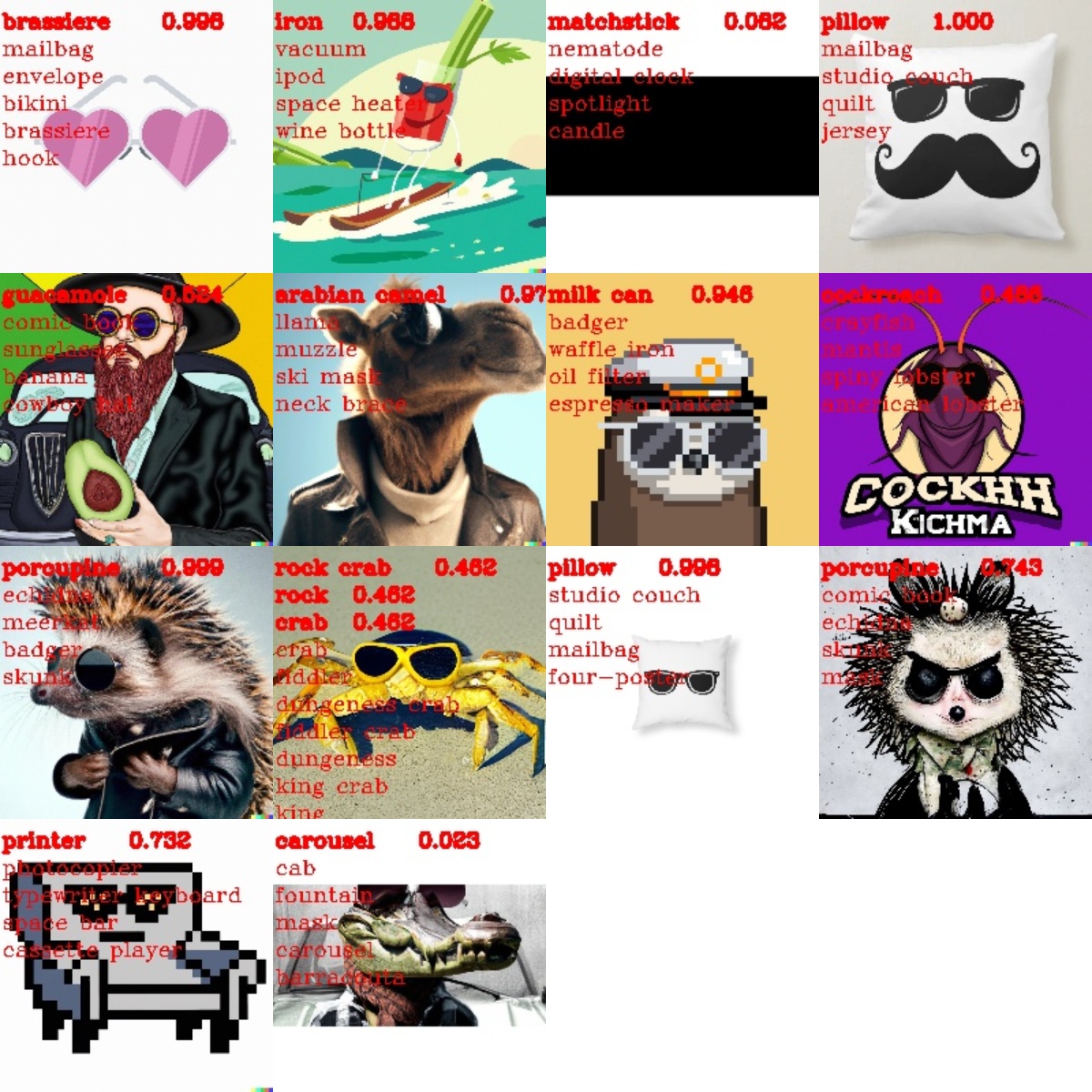}
\caption{Predictions of the resnext101\_32x8d\_wsl model over sunglass category.} 
\label{fig:Errors_sunglass}
\end{figure*}

\clearpage
\newpage
\section{Sample failure cases of best model (\texttt{resnext101\_32x8d\_ws}) over ImageNet-A dataset}
\label{appx:filures}

\begin{figure*}[htbp]
\centering 
 \includegraphics[width=1\linewidth]{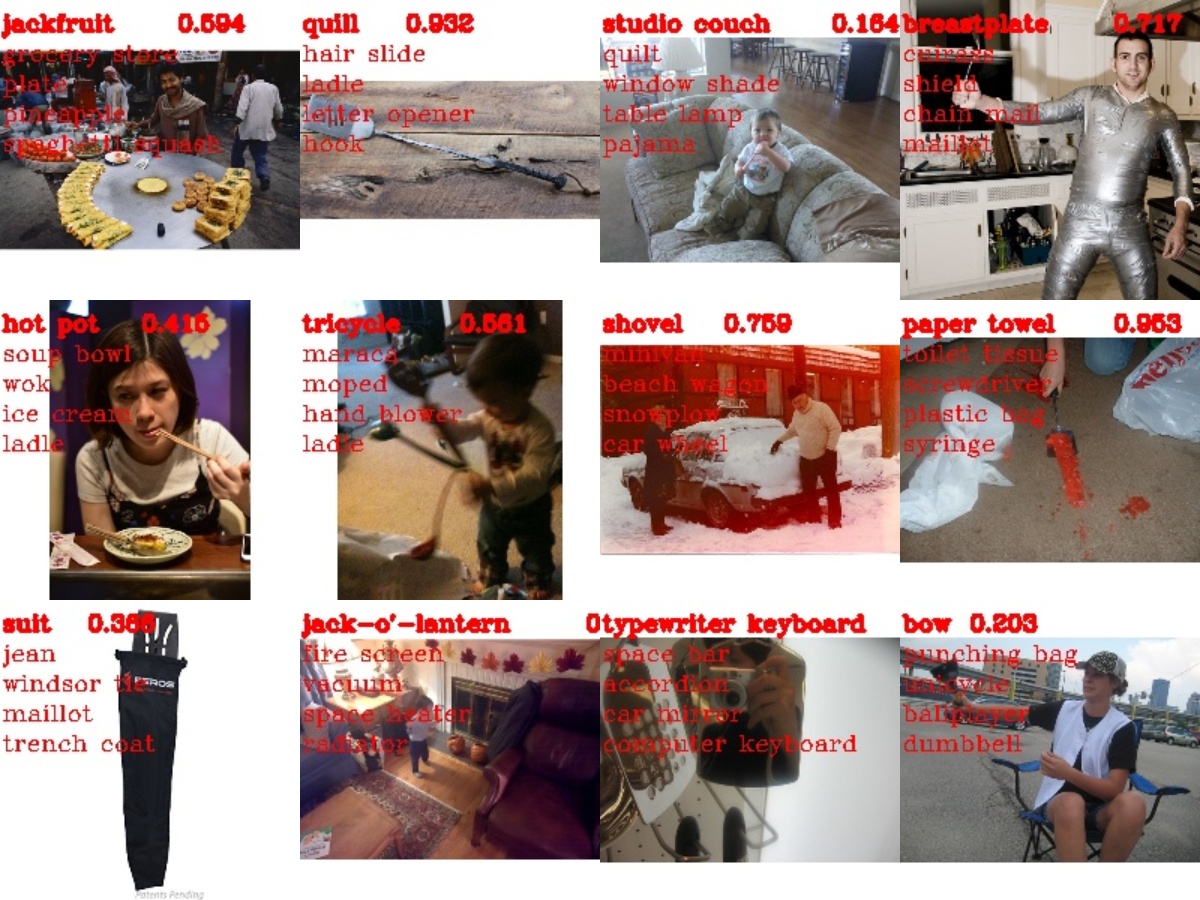}
\caption{Predictions of the resnext101\_32x8d\_wsl model over spatula category.} 
\end{figure*}

\clearpage
\newpage
\section{Sample failure cases of best model (\texttt{resnext101\_32x8d\_ws}) over SIsolated ImageNet-A dataset}
\begin{figure*}[htbp]
\centering 
 \includegraphics[width=1\linewidth]{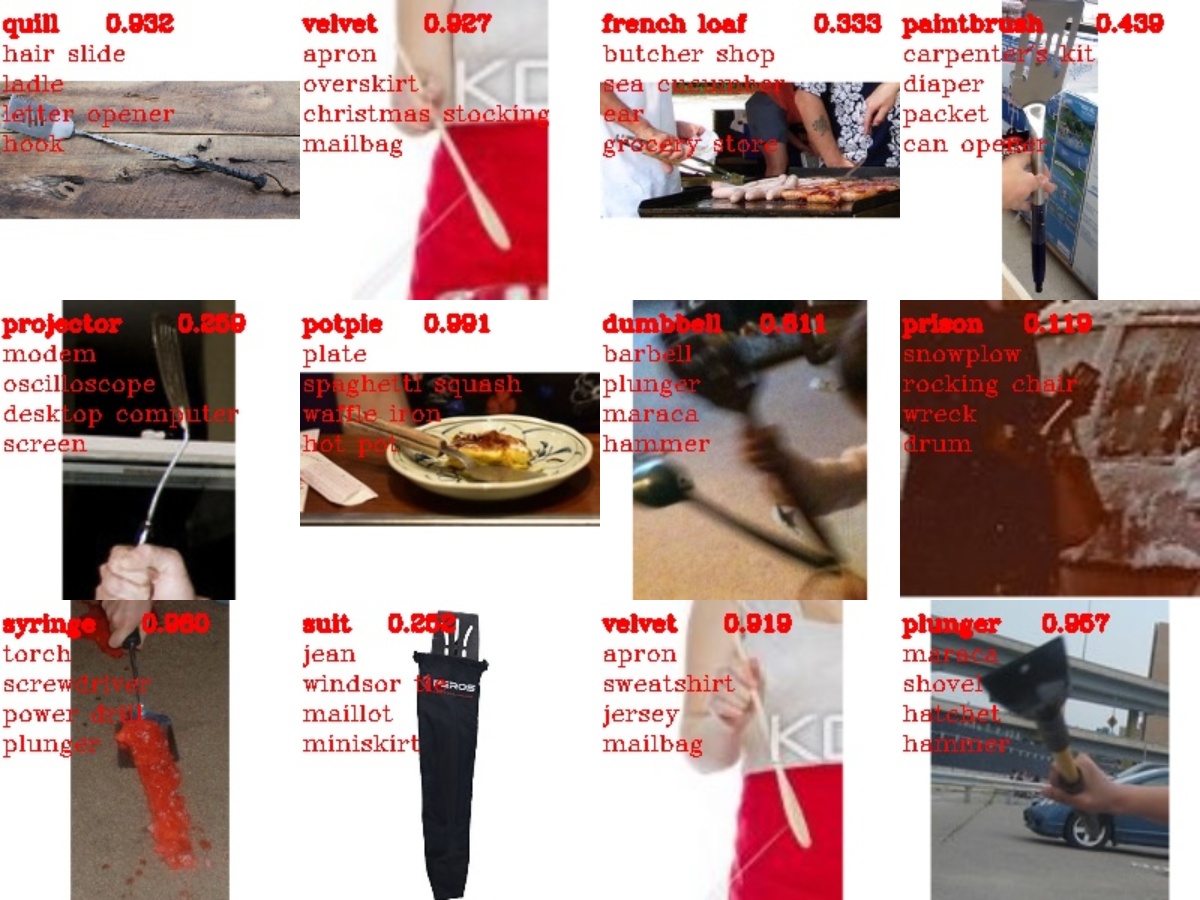}
\caption{Predictions of the resnext101\_32x8d\_wsl model over spatula category.} 
\end{figure*}

\clearpage
\newpage
\section{Predictions of APIs and models over face, person, cat, cow, giraffe, and car categories}
\label{appx:apis}

\begin{figure*}[htbp]
    \centering
    \includegraphics[width=\linewidth]{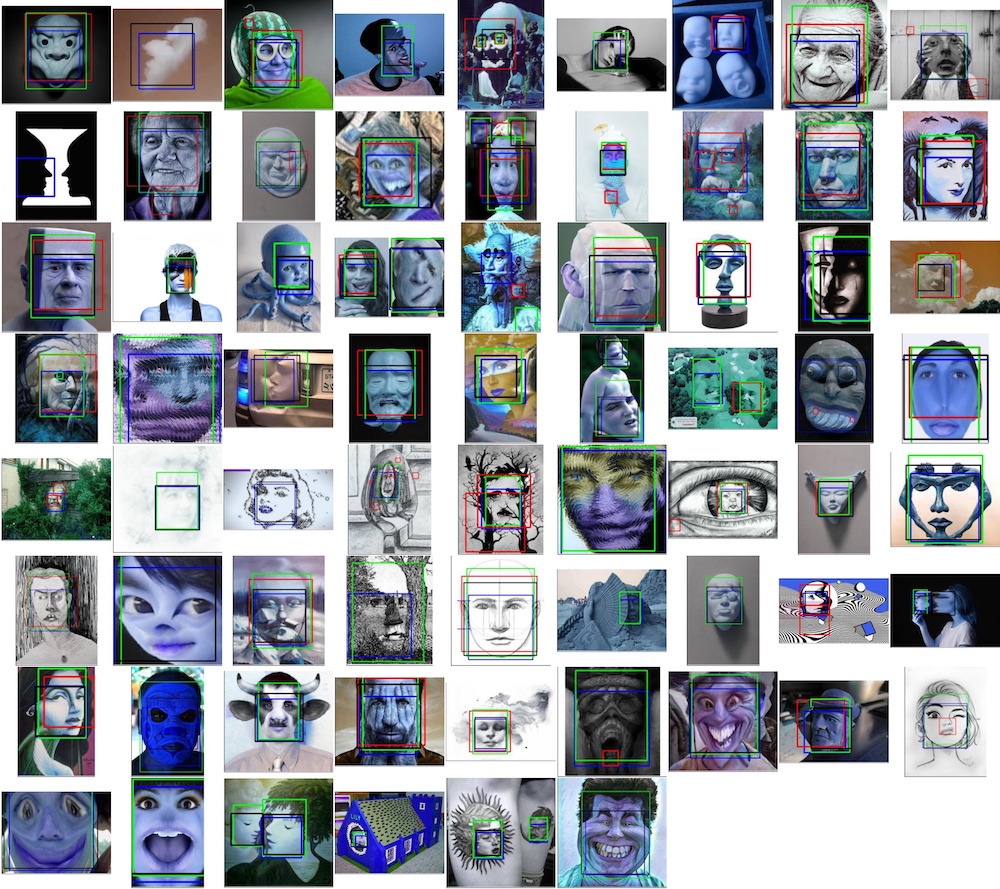}
    \caption{Sample face images along with predictions of OpenCV (Red), Microsoft API (Green), MEGVII Face++ API (Blue), and Google MediaPipe (Black) face detectors.}
\end{figure*}

\begin{figure*}[htbp]
    \centering
    \includegraphics[width=\linewidth]{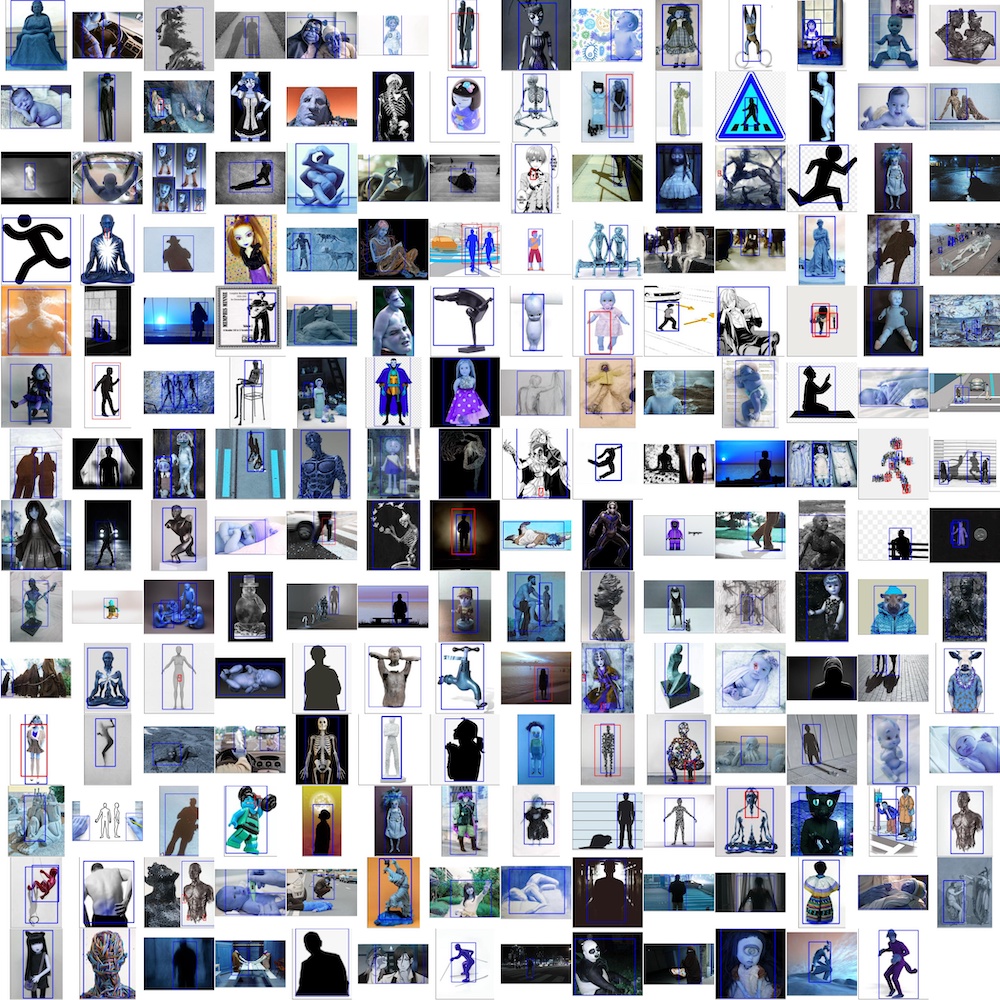}
    \caption{Sample person images along with predictions of the Microsoft API.}
\end{figure*}

\begin{figure*}[htbp]
    \centering
    \includegraphics[width=\linewidth]{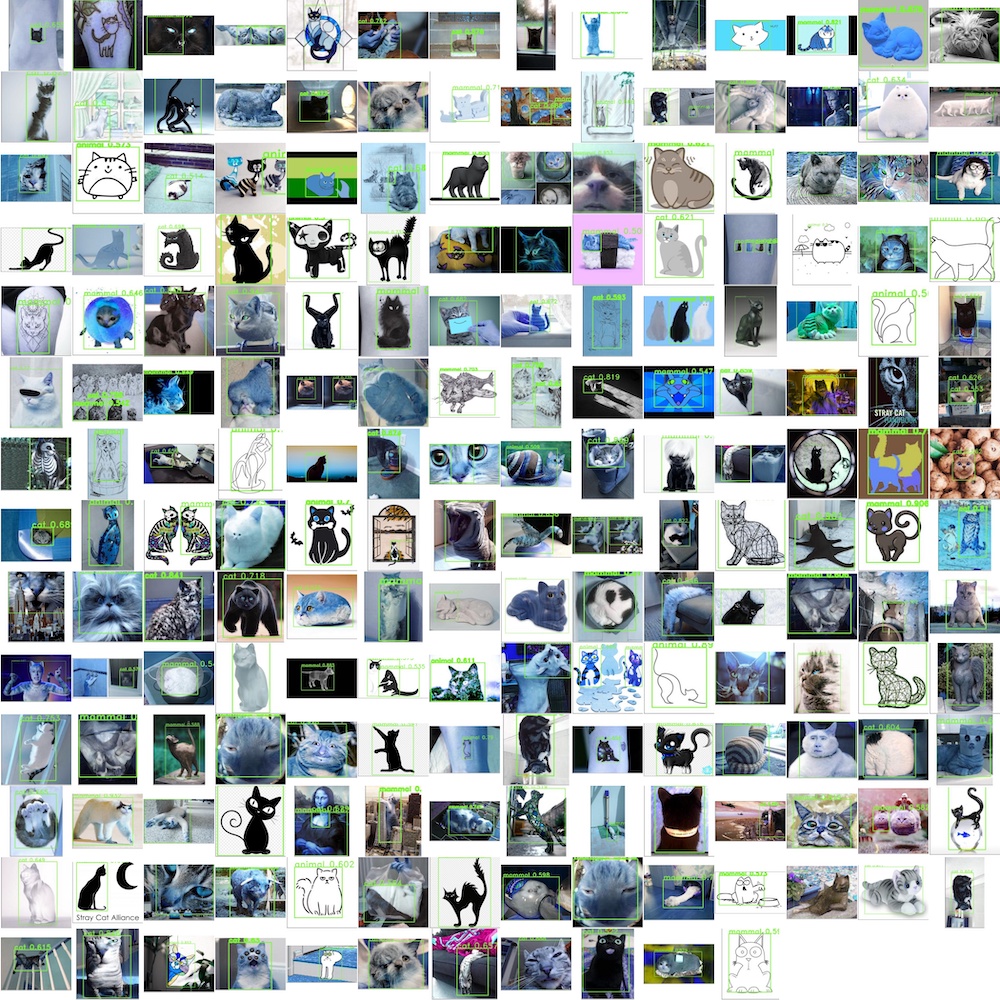}
    \caption{Sample cat images along with predictions of the Microsoft API.}
\end{figure*}







\clearpage

\section{Dataset License}
\label{appx:license}


D2O dataset is free to use only for research and academic purposes (not commercial). It is licensed under Creative Commons Attribution 4.0 with three additional clauses:

\begin{enumerate}
    \item D2O may never be used to tune the parameters of any model.
    \item The images containing people should not to be posted anywhere unless the people in the images are appropriately de-identified. Even in this case, written agreement from dataset creators is required. This is to check whether all the clauses are properly followed.
    
\end{enumerate}

To stop or limit the misuse of our D2O by bad actors, we have made a dataset request form\footnote{\url{https://bit.ly/3bDY0MS}}. We review the requests that we receive and allow
access for a legitimate use. The dataset we share contains images and questions is a zip file. The package also contains the detailed documentation with all relevant metadata specified to users.

\clearpage

\section{Sample generated images}
\label{appx:generated}

\begin{figure*}[htbp]
    \centering
    \includegraphics[width=.5\linewidth]{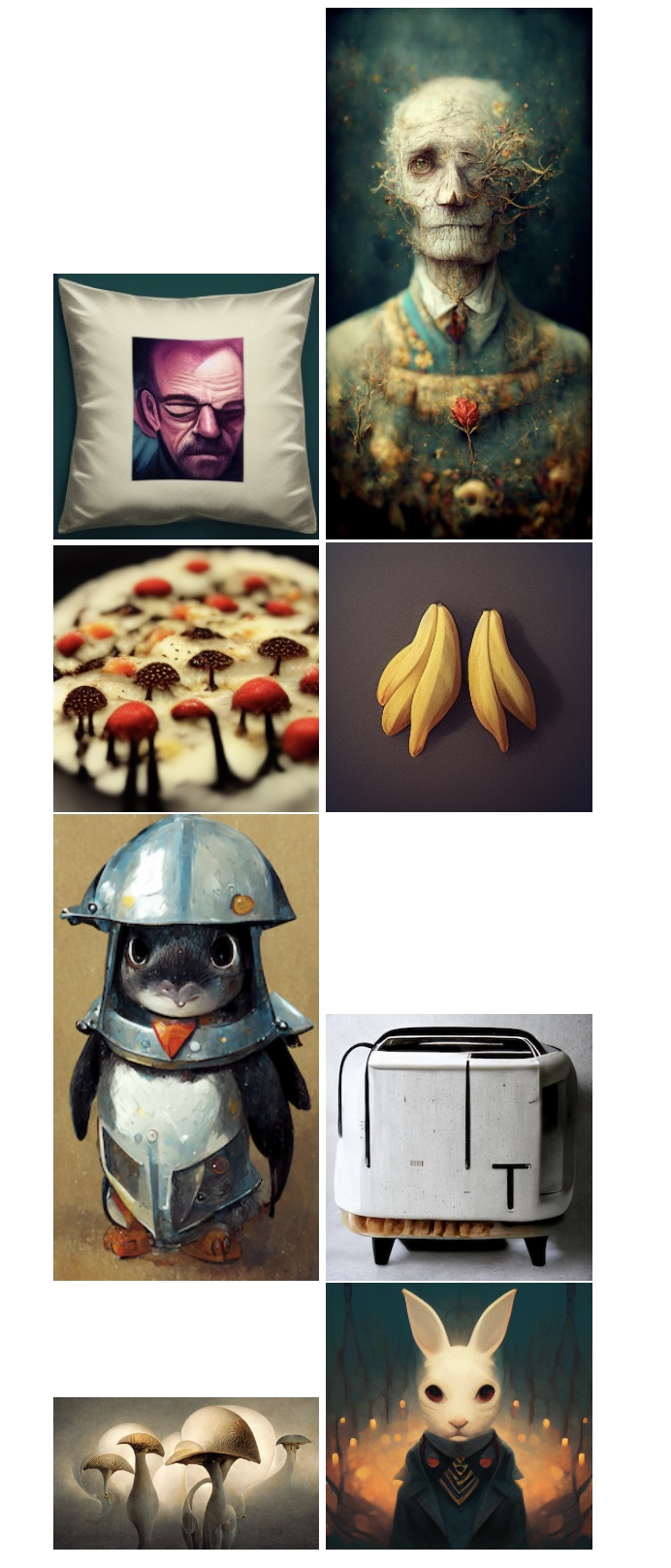}    
    \caption{Sample generated images by Midjourney from Pillow, Person, Pizza, Banana, Helmet, Toaster, Mushroom, and Rabbit classes.}
    \vspace{-100pt}
\end{figure*}

\end{document}